\documentclass[3p,review]{elsarticle}

\usepackage{natbib}
\usepackage{graphicx}
\usepackage{booktabs}
\usepackage{multirow}
\usepackage{subfig}
\usepackage{arydshln}
\usepackage{amsfonts}
\usepackage{amsmath}
\usepackage{algorithm}
\usepackage{algorithmic}
\usepackage{verbatim}
\usepackage{caption}
\usepackage{color,xcolor}
\usepackage{wrapfig}
\usepackage{graphicx}

\title{Deep Embedded Clustering with Distribution Consistency Preservation for Attributed Networks}

\author[1,2]{Yimei Zheng}
\ead{ymmzheng@bjtu.edu.cn}

\author[1,2]{Caiyan Jia \corref{cor1}}
\ead{cyjia@bjtu.edu.cn}

\author[1,2]{Jian Yu}
\ead{jianyu@bjtu.edu.cn}

\author[3]{Xuanya Li}
\ead{lixuanya@baidu.com}

\address[1]{School of Computer and Information Technology, Beijing Jiaotong University, Beijing, 100044, China}

\address[2]{Beijing Key Lab of Traffic Data Analysis and Mining, Beijing, 100044, China}

\address[3]{Baidu Online Network Technology (Beijing) Co., Ltd, Beijing, 100085, China}

\cortext[cor1]{Corresponding author.\\ 
E-mail addresses: ymmzheng@bjtu.edu.cn (Yimei Zheng), cyjia@bjtu.edu.cn (Caiyan Jia), jianyu@bjtu.edu.cn (Jian Yu), lixuanya@baidu.com (Xuanya Li).}

\begin{document}

\captionsetup[figure]{labelfont={bf},labelformat={default},labelsep=period,name={Fig.}}

\begin{frontmatter}

\begin{abstract}
  Many complex systems in the real world can be characterized by attributed networks. 
  To mine the potential information in these networks, deep embedded clustering, which obtains node representations and clusters simultaneously, has been paid much attention in recent years.
  Under the assumption of consistency for data in different views, the cluster structure of network topology and that of node attributes should be consistent for an attributed network. However, many existing methods ignore this property, even though they separately encode node representations from network topology and node attributes meanwhile clustering nodes on representation vectors learnt from one of the views. 
  Therefore, in this study, we propose an end-to-end deep embedded clustering model for attributed networks. It utilizes graph autoencoder and node attribute autoencoder to respectively learn node representations and cluster assignments. 
  In addition, a distribution consistency constraint is introduced to maintain the latent consistency of cluster distributions of two views. 
  Extensive experiments on several datasets demonstrate that the proposed model achieves significantly better or competitive performance compared with the state-of-the-art methods. The source code can be found at https://github.com/Zhengymm/DCP.
\end{abstract}

\begin{keyword}
Deep embedded clustering, Autoencoder, Graph autoencoder, Node representation learning, Cluster distribution consistency
\end{keyword}

\end{frontmatter}


\section{Introduction}

Many real-world systems can be modeled as networks, such as social networks, academic citation networks, and protein-protein interaction networks, where nodes denote objects and links denote pairs of relations between nodes. 
In addition to node connectivity information, nodes are often associated with attributes characterizing their own properties. Both kinds of information can implicitly reflect and affect the structures of networks. We generally call this kind of networks ``attributed networks''. For example, an academic citation network can be viewed as a representative attributed network. Each paper is a node of the network, and the citation relationship between a pair of papers constitutes edges in the network. The attributes of each paper are usually a bag-of-words characterizing the paper.
In recent years, there are more and more researches on attributed networks analysis, such as node classification, node clustering, and even recommendation \cite{GCN, DAEGC, lightGCN}. 

As the saying goes, `Birds of a feather flock together', therefore, clusters are important components in these attributed networks. 
In general, clustering aims to partition nodes in networks into disjoint clusters. In this way, nodes in the same cluster are not only more densely connected than nodes outside the cluster, but also the attribute similarity of a pair of nodes in the cluster is greater than that of a node pair in different clusters.
Therefore, identifying these clusters is an effective means to understand the underlying functions of systems that attributed networks represent.

Clustering has been studied extensively in data mining and machine learning fields over the past decades. 
Furthermore, with the development of deep learning in recent years, many deep models \cite{DEC, IDEC, s2ConvSCN} have been developed to promote the capacities of traditional clustering methods via neural networks. Since deep learning models \cite{DAEGC, MGAE} have the advantage of leveraging high-dimensional nonlinear features and high-relational features from nodes, they are becoming more and more resilient to high sparsity networks and attract considerable attention for node clustering, especially for large-scale networks.

The existing deep models for attributed network clustering can be roughly divided into two categories. The first one utilizes a network embedding method to obtain low-dimensional representation vectors with deep learning techniques \cite{DANE, ARGAE, AGE}, then adopts a typical clustering method, such as $K$-means \cite{kmeans} or spectral clustering \cite{sc}, to obtain node clusters. 
We call this kind of methods ``two-stage methods". 
However, they highly depend on the representation learning ability of deep models, and the learned representation may not be good for the task of node clustering, although they can also be directly used for other downstream tasks such as node classification and link prediction \cite{DANE, ARGAE}. In contrast, the second class of deep embedded clustering (DEC) methods directly integrate the objectives of clustering and representation learning into a unified framework \cite{DAEGC, SDCN, AANE}.
We call this kind of methods ``one-stage methods". Through joint optimization, DEC models are able to simultaneously learn node representations and the corresponding cluster structure, thereby gaining benefits and promoting each other.
The existing studies indicate that one-stage end-to-end methods for node clustering are better than two-stage methods \cite{DAEGC, SDCN}.

Recently, various DEC models \cite{DEC, IDEC, DBC} have been developed and arouse a lot of interest among researchers. However, most of the existing DEC models are designed to process image data, which can not be used to process attributed networks directly. Moreover, a given image dataset can be transformed into an attributed network by constructing a $k$NN ($k$-nearest neighbor) graph for obtaining global structure information, while keeping the image features as node attributes. Thus, it is necessary to further develop DEC models to capture cluster structure implied in network topology and node attributes while obtaining node representations for an attributed network. 

Since graph convolutional networks (GCN) provide a way to combine link structures and node features smoothly \cite{GCN}, some recent DEC models employ GCN as the backbone to make up for the deficiency of the previous methods that ignore topology information and achieve better clustering performance for attributed networks \cite{DAEGC, SDCN, AANE,NEC}.
DAEGC \cite{DAEGC} utilizes graph attention network (GAT) \cite{GAT} to form an autoencoder deep clustering framework, but lacks the reconstruction of node attributes which is a useful technique in self-supervised learning. SDCN \cite{SDCN} fuses autoencoder and GCN to learn node representations and hidden cluster distributions in a self-supervised way. However, it ignores to reconstruct the network topology.
We believe that reconstructing the original network (including both topology and node attributes) will be good for learning node representations. 
Meanwhile, it is worth investigating how to project the hidden vectors learned by the network into cluster distributions and maintain the actual cluster structure.
Corresponding to the consistency of multi-view clustering, the cluster structure of network topology and that of node attributes should be consistent for an attributed network. We suggest that the consistency constraint of cluster distributions obtained in the DEC process from different views is more robust than (or can be the substitute for) fusing two representation vectors before clustering. 

Based on these observations, in this study, we propose an end-to-end deep embedded clustering model named DCP-DEC (\underline{D}istribution \underline{C}onsistency \underline{P}reserving \underline{D}eep \underline{E}mbedded \underline{C}lustering) for attributed networks.
This model aims at utilizing network structure and node attributes to learn more comprehensive node representations and simultaneously achieve better clustering performance.
In the DCP-DEC model, we reconstruct a given network with a graph autoencoder (GAE) model where the backbone is still GCN. We keep DNN autoencoder in \cite{SDCN} to encode and decode node attributes. 
Once we obtain the node representation vectors from the above two autoencoders, we use them to obtain soft cluster assignments. Meanwhile, the KL-based clustering losses are then formed from two views in the means of self-supervised deep embedded clustering mechanism for network topology and node attributes.
For maintaining the latent distribution consistency of the obtained two cluster distributions, we further introduce a consistency constraint to the proposed model. The experimental studies demonstrate that the proposed model DCP-DEC is able to achieve our goal.

The main contributions of this study are summarized as follows.
\begin{itemize}
	\item We propose an end-to-end deep embedded clustering model for attributed networks, which exploits GAE and AE to learn node representations and cluster assignments simultaneously from network topology and node attributes.
	\item We further introduce a distribution consistency constraint to maintain the latent consistency of the two cluster assignments.
	\item Extensive experiments are carefully designed and conducted on various datasets, and the results demonstrate that our DCP-DEC model is highly competitive and even better than the SOTA methods.
\end{itemize}

The rest of this study is organized as follows. In section 2, we introduce the related works. The proposed deep embedded clustering model is presented in Section 3. In Section 4, we carefully analyze the performance of our model on several datasets compared with the state-of-the-art methods. Finally, the conclusion is drawn in section 5.

\section{Related works}
\subsection{Attributed network embedding}

Since attributed networks contain richer available information than pure networks, many attributed network embedding methods have been proposed \cite{NESurvey,NRLSurvey,GESurvey}. They utilize both information from topological structure and node attributes to learn more discriminative representations while capturing properties hidden in attributed networks.

One type of the representative attributed network embedding method is based on random walk. TriDNR \cite{TriDNR} exploits the inter-node relationship of a given node in a random walk sequence, and also captures the node-word correlation and label-word correspondence. Since \cite{GeneralizeRW} introduces the attributed random walk framework, it serves as a basis for generalizing many existing methods, such as DeepWalk \cite{DeepWalk} and node2vec \cite{node2vec}, to attributed network embedding.
The model proposed in \cite{IANRW} proposes a biased random walk between topology neighbors and attribute neighbors to get node representations.

In recent years, self-supervised deep models such as autoencoders and contrastive learning models \cite{SSL-CL} have played an important role in attributed network embedding tasks.
For instance, DANE \cite{DANE} employs an autoencoder to capture the non-linearity and proximity of link structures and node attributes. Furthermore, GAE/VGAE (Variational GAE) \cite{GAE} extends AE/VAE (Variational AE) \cite{VAE} to attributed networks to learn node representations using GCN as the encoder, and reconstructs the adjacency relations of nodes.
Moreover, recent breakthroughs in contrastive learning shed light on the potential of discriminative models for representation learning.
GraphCL \cite{graphcl} learns node embeddings by maximizing the similarity between the representations of two randomly perturbed versions of the intrinsic features and link structures of the same node’s local sub-graph. 
GRACE \cite{GRACE} generates two views by jointly corrupting both topological relations and node attributes, and learns node representations by maximizing the agreement of these two views at the node level.

Except for the self-supervised learning framework mentioned above, GCN-based encoding models, which take both network topology and node attributes into consideration, are conveniently used in attributed network embedding.
GraphSAGE \cite{GraphSAGE} leverages a set of aggregator functions that learn to aggregate node attribute information from a node’s local neighborhood to efficiently generate the embedding of the node.
AGE \cite{AGE} first applies a carefully-designed Laplacian smoothing filter, then employs an adaptive graph encoder that iteratively strengthens the filtered features to get better node embeddings in attributed networks. 

All these attributed network embedding methods devote to getting powerful node representations, then other downstream tasks such as node clustering, node classification, and link prediction can be carried out directly on the learned representation vectors. However, the learned node vectors may not be the best for the task of node clustering. It is the concern of this study.

\subsection{Deep embedded clustering}
Deep embedded clustering models gradually come into the field because they can jointly optimize node clustering and representations. 
As a pioneer, the DEC method \cite{DEC} trains an autoencoder to learn representation vectors and defines a clustering loss based on them to obtain clusters. It provides a way to learn node representations and cluster assignments simultaneously with a shared encoder.
IDEC \cite{IDEC} further extends the DEC method \cite{DEC} to integrate the reconstruction loss of AE and the clustering loss of DEC into a unified objective, training the model in an end-to-end framework. 
Subsequently, a number of approaches for deep spectral clustering and deep subspace clustering inspired by them also begin to emerge \cite{s2ConvSCN,DSpectralC,DSubspaceC}.

However, the above methods cannot be used to process attributed networks directly since they ignore the link structures of data. To solve this limitation, a few methods which inherit the advantages of GCN-based models have been designed and used for DEC tasks for attributed networks. For example, 
DAEGC \cite{DAEGC} employs GAT network \cite{GAT} as the encoder to rank the importance of attributed nodes within a neighborhood and learn node representations.
It is able to supervise the process of clustering to get clusters in the means of self-training for attributed networks. \cite{DNENC} extends the encoder of DAEGC to GCN to further aggregate the neighbor information of nodes in the network.
NEC \cite{NEC} introduces a relaxed soft modularity that can be optimized with the reconstruction loss of GAE. Furthermore, it combines them with the KL divergence based clustering loss, thus improving cluster assignments and feature representations in a self-learning manner.
The model proposed in \cite{EGAE_JOCAS} incorporates a joint clustering model into the graph encoder. It fuses relaxed $K$-means and spectral clustering to get clusters, and the adjacency is shared by both GAE and joint clustering. 
Besides, SDCN \cite{SDCN} first combines GCN and DNN autoencoder with a delivery operator and then designs a dual self-supervised mechanism for clustering to get the final cluster assignments through joint training. However, it directly projects the representation vectors of GCN into cluster distributions by a subsequent GCN layer without the real clustering process, thus may lead to poor clustering performance.
Based on SDCN, DFCN in \cite{DFCN} fuses two representation vectors learned from IGAE on an attributed graph and AE on node attributes, then applies triplet self-supervised clustering on the learned vectors to preserve the cluster structures in the embedding space. What's more, after the learning process of DFCN, $K$-means clustering is used again on the final fused representation vectors to get the cluster assignments. 
\cite{GC_VGE} combines a variational graph autoencoder for network embedding with a self-training mechanism into a unified framework, enabling the better clustering of networks.
Recently, O2MAC \cite{One2Multi} extends DEC models to multi-view attributed networks. It consists of one encoder and multiple decoders to learn the shared representations. Furthermore, a self-training clustering objective is also designed to optimize node embeddings and clusters simultaneously. In this study, we concentrate on this branch of research and intend to design a more powerful one-stage end-to-end model to learn node representations effectively and efficiently for the task of node clustering.

\section{Proposed methodology}

\subsection{Notations and problem definition}

In this study, an undirected attributed network is represented as a graph $G = (V; E; \mathbf{X})$, where $\vert V \vert=n$ consists a set of $n$ nodes $V =\{v_1,v_2,\ldots,v_n\}$, and $E$ represents a set of edges. The topological structure of $G$ is specified by an adjacency matrix $ \mathbf{A}\in \mathbb{R}^{n \times n}$. If there is an edge between node $v_i$ and $v_j$ (i.e., $ e_{ij} = (v_i, v_j) \in E$), then $a_{ij} = 1$, otherwise $a_{ij} = 0$. Furthermore, $ \mathbf{X}\in \mathbb{R}^{n \times m}$ represents the attribute matrix, where $m$ is the dimension of attributes. Each row of $\mathbf{X}$ describes the attributes of node $v_i$, which can be discrete binary values or continuous real-values. 

Given an attributed network $G$, deep embedded clustering aims to learn low-dimensional node representations while clustering the network into groups. 
The two tasks are expected to be optimized jointly in an end-to-end unified framework, and promote each other during the training process. 
Specifically, our purpose is to learn a map function $f:(\mathbf{A}, \mathbf{X}) \mapsto \mathbf{Z} \in \mathbb{R}^{n \times d}$, $z_i$ is the $i$-th row of $\mathbf{Z}$, which represents the latent representations of node $v_i$, and the dimension of $z_i$ is $d$, generally $d<m$ \& $d<n$. Meanwhile, we intent to partition the attributed network into $K$ disjoint groups (${G_1,G_2,\ldots,G_K}$) such that nodes in the same group are not only more densely connected than nodes outside the group, but also the attribute similarity of a pair of nodes in the group is greater than that of a node pair with only one in the group. 

\begin{figure*}[ht]
	\centering
	\includegraphics[width=\textwidth]{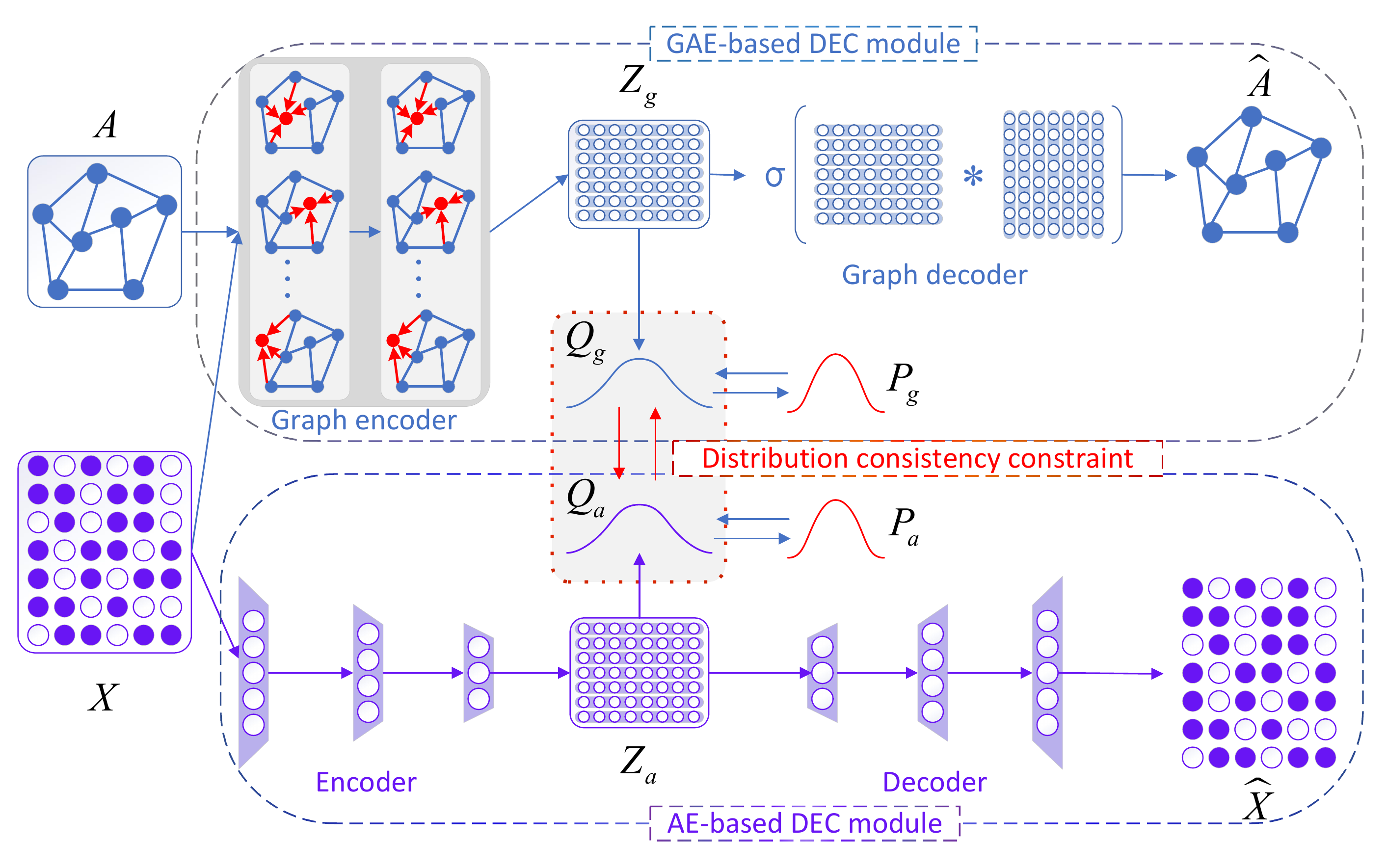}
	\caption{The framework of DCP-DEC model. The three dotted boxes correspond to GAE-based DEC module, AE-based DEC module, and distribution consistency constraint. $\mathbf{Q}_g$ and $\mathbf{Q}_a$ are soft cluster assignments designed for the former two parts, where $\mathbf{Q}_g$ is our final output for clustering.}
	\label{fig: framework}
\end{figure*}

\subsection{Overview of the framework}

In this section, we will introduce our end-to-end deep embedded clustering model DCP-DEC (\underline{D}istribution \underline{C}onsistency \underline{P}reserving \underline{D}eep \underline{E}mbedded \underline{C}lustering) for attributed networks.
Fig.~\ref{fig: framework} illustrates the framework of our proposed model. It mainly consists of three parts: autoencoder (AE) based DEC module for node attributes, graph autoencoder (GAE) based DEC module mainly for network topology, and distribution consistency constraint to balance and fuse the above two modules.

\begin{itemize}
	\item {\textbf{AE-based DEC module}}. This module adopts DNN to encode and decode node attributes to learn representations of nodes. With the obtained vectors, we get the soft cluster assignment for each node through Student's t-distribution and then correspondingly construct a target distribution.
	During the training, we minimize the reconstruction loss of the decoder and the clustering loss in a unified way, where the clustering loss is characterized by KL divergence between the target distribution and the soft assignment. 
	Therefore, in this step, we only use attribute information of nodes to get clustering results and node representations.
	\item {\textbf{GAE-based DEC module}}. In this module, we fuse network structure and node attributes together to learn latent representations using the GCN encoder, then reconstruct link connections via an inner product decoder to guide the learning process. As is well known, network topology is usually very sparse, so nodes attributes are very helpful to guide the reconstruction of network structures. 
	The clustering process is conducted in the same way as the above AE-based DEC module to get cluster assignments in this view. Obviously, in this module, we use network topology as the main information source and attribute information as the complementary.
	\item {\textbf{Distribution consistency constraint}}. From the above two modules, we have obtained two cluster assignments using different information and different learning processes. Based on the assumption of consistency for data in different views, we introduce a distribution consistency constraint to maintain the consistency of the cluster structure of node attributes and that of network topology (enhanced by the node attributes). By minimizing the KL divergence between two cluster distributions from the above two modules, we can obtain a more robust and smooth cluster structure, thus better fusing two different views for the task of attributed network clustering.
\end{itemize}

\subsection{DCP-DEC model}

\subsubsection{AE-based deep embedded clustering}

AE-based module only encodes node attributes to learn node representations, and uses a decoder to reconstruct the input. 
In this way, various autoencoders and their variants can be used, including variational autoencoder \cite{VAE}, denoising autoencoder \cite{DAE}, and stacked autoencoder \cite{CAE}, etc. 
In this study, we simply use a fully connected DNN as an encoder to map the attributes of each node to a nonlinear low-dimension latent representation vector. The encoder can be formulated in the following.
\begin{equation}
\mathbf{Z}^{(l)}_a = \phi(\mathbf{W}^{(l)}_e \mathbf{Z}^{(l-1)}_a + \mathbf{b}^{(l)}_e)
\end{equation}
where $l\in \{1,2,\ldots,L\}$ is the number of layers for DNN, $\mathbf{Z}^{(0)}_a = \mathbf{X}$ is the initial input of node attributes, $\mathbf{W}^{(l)}_e$ and $\mathbf{b}^{(l)}_e$ are weights and bias of the $l$-th layer in the encoder, and $\phi$ is the activation function, such as ReLU or Sigmoid.

Then a decoder follows after the encoder. It employs the representation of the last layer in the encoder to reconstruct the input attributes. Also, there are $L$ fully connected layers, which are formulated below.
\begin{equation}
\label{za}
\hat{\mathbf{Z}}^{(l)}_a = \phi(\mathbf{W}^{(l)}_d \hat{\mathbf{Z}}^{(l-1)}_a + \mathbf{b}^{(l)}_d)
\end{equation}
where $l\in \{1,2,\ldots,L\}$, $\mathbf{W}^{(l)}_d$ and $\mathbf{b}^{(l)}_d$ are weights and bias of the $l$-th layer in the decoder, respectively, and $\hat{\mathbf{Z}}^{(0)}_a=\mathbf{Z}^{(L)}_a$. The output of the decoder is the reconstructed attributes $\hat{\mathbf{X}}=\hat{\mathbf{Z}}^{(L)}_a$. Thus, the reconstruction loss of the autoencoder is defined as:
\begin{equation}
\label{l_are}
\mathcal{L}_{are} = \frac{1}{2N} {\Vert \mathbf{X} - \hat{\mathbf{X}} \Vert}^2_2
\end{equation}

Once we obtain the latent representation $\mathbf{Z}^{(L)}_a \in \mathbb{R}^{n \times d}$ from the attributes of nodes, we can utilize it to generate the soft cluster assignment $\mathbf{Q}_a \in \mathbb{R}^{n \times K}$, where $q^a_{ik}$  obtained by Eq.~\ref{qik_a} represents the probability that node $v_i$ is assigned into the cluster $k$. Since $q^a_{ik}$ measures the similarity between the node representation $z^a_i$ and clustering center $\mu^a_k$ through Student's t-distribution as a kernel, once nodes are closer to the cluster center, the corresponding soft assignments get high probability and are more likely to be trustworthy.
Then the target distribution $\mathbf{P}_a$ of nodes is successfully constructed through Eq.~\ref{pik_a}, which intends to put more emphasis on a data point that is assigned with high confidence to improve cluster purity. 
It is worth noticing that to get the initial centers $\mu^a_k$ ($k=1,2, \cdots, K$), we first pre-train the autoencoder through only minimizing the reconstruction loss to obtain the meaningful representation $\mathbf{Z}_a$. After that, we run $K$-means on the learned representations for initializing, and in the following iterative training, the $K$-means clustering is never used again.
\begin{equation}
\label{qik_a}
q^a_{ik} = \frac{(1 + \Vert z^a_i - \mu^a_k \Vert ^2)^{-1}}{\sum_j(1 + \Vert z^a_i - \mu^a_j \Vert ^2)^{-1}}
\end{equation}
\begin{equation}
\label{pik_a}
p^a_{ik} = \frac{{q^a_{ik}}^2 / \sum_i q^a_{ik}}{\sum_{j} ({q^a_{ij}}^2 / \sum_i q^a_{ij})}
\end{equation}

Following the previous studies \cite{DAEGC,DEC,IDEC,SDCN}, we aim to force the current assignment $\mathbf{Q}_a$ to approach the target distribution $\mathbf{P}_a$, such that the final clusters can be obtained through jointly training autoencoder and cluster assignments.
The clustering loss is defined according to the KL divergence between two distributions as below.
\begin{equation}
\label{kl_ae}
\mathcal{L}_{aKL} = KL(\mathbf{P}_a \Vert \mathbf{Q}_a) = \sum_i \sum_k p^a_{ik} \log \frac{p^a_{ik}}{q^a_{ik}}
\end{equation} 

In order to get reliable clusters, we minimize Eq.~\ref{kl_ae}, when $\mathbf{Q}_a$ reaches to idempotent, it closes to 0. At this condition, the cluster structure is clear and the entropy of the cluster distribution is minimal.
This strategy is also called ``self-supervised", which is the main idea of deep embedded clustering \cite{DEC}.

In brief, AE-based deep embedded clustering module jointly optimizes node representations and the cluster structure using only node attributes. 
The whole objective can be formulated as:
\begin{equation}
\label{l_ae}
\mathcal{L}_{AE} = \mathcal{L}_{are} + \alpha \ast \mathcal{L}_{aKL}
\end{equation}
where $\alpha \ge 0$ is a control coefficient that balances the reconstruction loss and the clustering loss.

\subsubsection{GAE-based deep embedded clustering}

As for attributed networks, not only node attributes can reflect the properties of clusters, but also the network topology indicates the relationships between nodes to form clusters. 
Moreover, previous study \cite{DANE_att} has shown that if only use link connections to partition networks, the obtained cluster structure may not be good since the sparseness of network topology in the real world.
Hence, the GAE-based module exploits both kinds of information to achieve deep embedded clustering. It complements the AE-based module from another view and is likely to help us obtain a more robust and reliable final cluster distribution, meanwhile getting powerful node representations.

Specifically, we employ two-layer GCN as an encoder to learn node representations since it has a powerful ability to retain the information in an attributed network \cite{AM-GCN, over-smooth_GCN}. In general, the layer-wise transformation of the encoder can be expressed as below.
\begin{equation}
\label{zg}
\mathbf{Z}^{(l)}_g = \phi(\tilde{\mathbf{D}}^{-\frac{1}{2}} \tilde{\mathbf{A}} \tilde{\mathbf{D}}^{-\frac{1}{2}} \mathbf{Z}^{(l-1)}_g \mathbf{W})
\end{equation} 
where $\tilde{\mathbf{A}} = \mathbf{A} + \mathbf{I}$, represents the topology of a network with a self-loop on each node, $\mathbf{I}$ and $\tilde{\mathbf{D}}$ are the identity diagonal matrix and degree matrix of nodes respectively, where $\tilde{D}_{ii} = \sum_j \tilde{a}_{ij}$. 
In addition, $\mathbf{\mathbf{Z}}^{(0)}_g =\mathbf{X}$ means that node attributes are used as the initial representation vector for each node, $\mathbf{W}$ is the trainable weight matrix, $\phi$ is the ReLu activation function.

In order to preserve the original network structure and make the learned representations more discriminative, we also introduce a decoder to guide the learning process. 
In the literature, there are various kinds of decoders, which reconstruct either the graph structure \cite{ARGAE, GAE}, node attributes \cite{MGAE, GALA}, or both of them \cite{GUCD, AN2VEC}. 
As GCN views network topology as the main information and nodes with similar attributes tend to link together, we use the inner product decoder to reconstruct the relationships between nodes.
This works as a counterpart to the former AE-based module and is formulated in the following.
\begin{equation}
\hat{\mathbf{A}} = Sigmoid(\mathbf{Z}^{(L)}_g {\mathbf{Z}^{(L)}_g}^{T})
\end{equation}

Similarly, by minimizing the reconstruction error of the GAE in Eq.~\ref{l_gre}, we can get another node representation vector $\mathbf{Z}_g^{(L)}$.
\begin{equation}
\label{l_gre}
\mathcal{L}_{gre} = \frac{1}{2N} {\Vert \mathbf{A} - \hat{\mathbf{A}} \Vert}^2_2
\end{equation}

After getting node representations from GAE, we can perform the same clustering operation as the AE-based DEC module of node attributes.
Similar to Eqs.~\ref{qik_a} and \ref{pik_a}, we utilize $\mathbf{Z}^{(L)}_g$ to generate the soft cluster assignment $\mathbf{Q}_g$ and target distribution $\mathbf{P}_g$, where the initial cluster centers $\mu^g_k$ ($k=1,2, \cdots, K$) are obtained by running $K$-means on the representation of the pre-trained GAE. The pre-training process is the same for AE and GAE models, we will describe it in detail in the experiments section.
Then, we iteratively update node representations and nodes' clusters by learning from the high confidence assignments through optimizing the following clustering loss.
\begin{equation}
\mathcal{L}_{gKL} = KL(\mathbf{P}_g \Vert \mathbf{Q}_g) = \sum_i \sum_k p^g_{ik} \log \frac{p^g_{ik}}{q^g_{ik}}
\end{equation} 

The same as before, in order to learn better node representations for the task of clustering, we combine the reconstruction loss and the clustering loss together for this GAE-based module. The objective function in this part is defined as below.
\begin{equation}
\label{l_gae}
\mathcal{L}_{GAE} = \mathcal{L}_{gre} + \beta \ast \mathcal{L}_{gKL}
\end{equation}
where $\beta \ge 0$ is a factor that controls the balance between the two losses. Therefore, we can refine cluster assignments and simultaneously obtain node representations by jointly training. 

Although the AE-based and GAE-based DEC modules apply the same idea and look similar, the information they have contained is different.
Obviously, the major difference between these two modules is that AE-based module takes node attributes as its information sources, while GAE-based module takes network topology as the major information and node attributes as auxiliary.  
Note that if a GCN model has many convolutional layers, the output features may be over-smoothed and vertices from different clusters may become indistinguishable, thus the clustering performance will be affected \cite{SDCN,over-smooth_GCN}.
Besides, multiple layers of GCN will increase the computational complexity. 
However, shallow GCNs with only two or three layers may not have the ability to fully mine the information hidden in an attributed network.
Fortunately, it has been proved that fusing other information such as node attributes to GCN is able to strengthen GCN's learning ability. 
Therefore, in this study, we use shallow GCN and take node attributes as complementary, such that our model is able to efficiently learn effective node representations and cluster assignments.

\subsubsection{Distribution consistency constraint}

As mentioned above, we can obtain two node representations and cluster distributions from two different deep embedded clustering modules, AE-based DEC module and GAE-based DEC module.
Under the assumption of cluster consistency for data objects in different views, the cluster structure of network topology and that of node attributes should be consistent for an attributed network. 
Therefore, we propose a clustering distribution consistency constraint to make the assignments of clusters learned from the above two DEC modules as consistent as possible. 
We believe that this constraint is more robust to get good clusters or can be the substitute for fusing two representation vectors before clustering. 

We employ KL divergence to measure the difference between two soft assignments $\mathbf{Q}_g$ and $\mathbf{Q}_a$ in Eq.~\ref{l_con}.
Since $\mathbf{Q}_g$ is obtained from the GAE module, which exploits both topology and node attributes, we believe that $\mathbf{Q}_g$ contains richer information than $\mathbf{Q}_a$ that only considers attributes. We utilize $\mathbf{Q}_g$ as the guidance to form the KL constraint. That is to say, we encourage the cluster distribution learned from the AE-based module to match that learned from the GAE-based module. 
\begin{equation}
\label{l_con}
\mathcal{L}_{con} = KL(\mathbf{Q}_g \Vert \mathbf{Q}_a) = \sum_i \sum_k q^g_{ik} \log \frac{q^g_{ik}}{q^a_{ik}}
\end{equation}

During minimizing the KL divergence between $\mathbf{Q}_g$ and $\mathbf{Q}_a$, the consistency constraint leads the two cluster distributions to be similar and makes the model satisfy the cluster consistency assumption.
In this way, the representations of nodes learned from two DEC modules are updated iteratively under the guidance of this constraint, forming a unified end-to-end framework. 

\subsubsection{Overall objective function and optimization}

To sum up, the above three parts cooperate with each other to jointly optimize the process of learning node representations and clustering nodes in networks. 
The overall loss function of the DCP-DEC model is defined as:
\begin{equation}
\label{loss}
\mathcal{L} = \mathcal{L}_{AE} + \mathcal{L}_{GAE} + \gamma \ast \mathcal{L}_{con}
\end{equation}

These three items correspond to the three parts mentioned above. The first is the loss of AE-based module, the second is that of GAE-based module and the last is the distribution consistency constraint. 
In addition, $\gamma \ge 0$ is the coefficient controlling the weight of consistency constraint to balance the whole loss during training.
We expect to minimize Eq.~\ref{loss} of a given attributed network via stochastic gradient descent in back-propagation, and obtain better cluster assignments meanwhile learning discriminative node representations.
The whole learning process of the proposed DCP-DEC model is shown in Algorithm~\ref{alg:algorithm}.

It is worthy of mentioning that the newly proposed model DCP-DEC outputs the final clustering result directly from the cluster assignment $\mathbf{Q}_g$, which matches our one-stage strategy. 
Namely, the cluster $c_i$ for node $v_i$ can be estimated as follows. 
\begin{equation}
c_i = \mathop{\arg \max}\limits_k q^g_{ik}
\end{equation}

\begin{algorithm}[ht]
	\renewcommand{\algorithmicrequire}{\textbf{Input:}}
	\renewcommand{\algorithmicensure}{\textbf{Output:}}
	\caption{Distribution Consistency Preserving Deep Embedded Clustering}
	\label{alg:algorithm}
	\begin{algorithmic}[1]
		\REQUIRE An attributed network $G$ with adjacent matrix $A$ and attribute matrix $X$, the number of clusters $K$, and the maximum number of iterations $MaxIter$.
		\ENSURE Clustering results $C$.
		\STATE Train AE and GAE by minimizing Eqs.~\ref{l_are} and \ref{l_gre} to get the initial parameters of AE and GAE and obtain their initial representations $\mathbf{Z}_a$ and $\mathbf{Z}_g$; \COMMENT{Pre-training}
		\STATE Run $K$-means on $\mathbf{Z}_a$ and $\mathbf{Z}_g$ to get initial cluster centers $\mu^a$ and $\mu^g$, respectively.  \COMMENT{Pre-training}
		\FOR [Training] {$iter \in 
		\{0,1, \cdots, MaxIter\}$}  
		\STATE Learn node representations $\mathbf{Z}_a$ and $\mathbf{Z}_g$ via Eqs.~\ref{za} and \ref{zg};
		\STATE Compute soft assignments $\mathbf{Q}_a$ and $\mathbf{Q}_g$ of two modules;
		\STATE Compute target distributions $\mathbf{P}_a$ and $\mathbf{P}_g$, correspondingly;
		\STATE Calculate loss of AE-based and GAE-based modules $\mathcal{L}_{AE}$ and $\mathcal{L}_{GAE}$ via Eqs.~\ref{l_ae} and \ref{l_gae};
		\STATE Calculate distribution consistency constraint $\mathcal{L}_{con}$ via Eq.~\ref{l_con};
		\STATE Update the whole model by minimizing Eq.~\ref{loss};
		\ENDFOR
		\RETURN $C$, where $c_i = \mathop{\arg \max}\limits_k q^g_{ik} (i=1,\cdots, n; k=1, \cdots, K)$.
	\end{algorithmic}
\end{algorithm}

\section{EXPERIMENTS}

\subsection{Experiment settings}

\subsubsection{Datasets}

We validate the proposed model on eight real-world datasets and several artificial attributed networks, the detail information of these networks are summarized in Table~\ref{tab:datasets} and Table~\ref{tab:LFR}, respectively. 

The real-world datasets have five attributed networks, including three popular citation networks\footnote{https://linqs.soe.ucsc.edu/data} (i.e., Citeseer, Cora, and PubMed, where nodes represent documents and each document is described by a 0/1-value word vector for Citeseer and Cora and a TF/IDF weighted word vector for PubMed), 
one paper network (i.e., ACM\footnote{https://dl.acm.org/}, where there is an edge between two papers if they are written by the same author and the node features are represented by the bag-of-keywords in each paper), 
and one author cooperation network (i.e., DBLP\footnote{https://dblp.uni-trier.de/}, where the features of each author are characterized by the bag-of-keywords related to this author). 
Besides, the real-world datasets contain three non-graph datasets, including USPS \cite{USPS}, HHAR \cite{HHAR}, and Reuters \cite{Reuters}. 
In detail, USPS is a gray-scale handwritten digit image dataset, where the features of each image are described by the gray value of pixel points in the image. HHAR contains sensor records from smart phones and smart watches, where the features of each record are composed of commonly used time and frequency features. Reuters is a text dataset containing around 810,000 English news stories. As usual \cite{DEC}, for the task of text clustering, we select 4 root categories: corporate/industrial, government/social, markets and economics, and sample a random subset with 10,000 stories where each story is characterized by 2,000 most frequent words with their TF/IDF values. For each of these three non-graph datasets, we construct a $k$NN graph, and combine its original features to form the corresponding attributed network. Following the previous studies \cite{SDCN, DFCN}, we choose $k$
with the best performance for each dataset. The selected number of edges for the constructed $k$NN graph of each dataset is also shown in Table~\ref{tab:datasets}.

\begin{table}[ht]
\begin{center}
    \caption{Statistics of real-world datasets.}
	\label{tab:datasets}
	\begin{tabular}{cccccc}
		\toprule
		Dataset & Nodes & Edges & Attributes & Clusters & $\# k$ \\
		\midrule
		Citeseer & 3,327  & 4,732   & 3,703 & 6   & -  \\
		Cora     & 2,708  & 5,429   & 1,433 & 7   & -  \\
		PubMed   & 19,717 & 44,338  & 500   & 3   & -  \\
		ACM 	 & 3,025  & 26,256  & 1,870 & 3   & -  \\
		DBLP   	 & 4,057  & 7,056   & 334   & 4   & -  \\
		USPS     & 9,298  & 46,490  & 256   & 10  & 5  \\
		HHAR	 & 10,299 & 61,794  & 561   & 6   & 6  \\
		Reuters	 & 10,000 & 100,000 & 2,000 & 4   & 10 \\
		\bottomrule
	\end{tabular}
\end{center}
\end{table}

The artificial attributed networks are mainly generated by LFR benchmark \cite{LFR} with the settings showed in Table~\ref{tab:LFR}, where $N$ is the number of nodes, $\mu$ is the mixing parameter, $\langle k \rangle$ is the average degree of nodes, $k_{max}$ is the maximum degree of nodes, $C_{min}$ and $C_{max}$ are the minimum and maximum cluster sizes, $\gamma$ and $\beta$ are exponents of the power-law distribution of node degree and cluster size, respectively. 
Among these, $\mu$ is designed to control the clearness of cluster structure in a network. Each node shares a fraction $1 - \mu$ of its links with other nodes in its cluster and a fraction $\mu$ of its links with the other nodes in the network. Thus, the smaller $\mu$, the clearer cluster structure will be. 
Since previous studies have shown that almost all methods can achieve very good performance when network structure is clear, we only focus on the scenarios where the cluster structure is unclear, i.e., $\mu =\{0.6, 0.7, 0.8\}$. In addition, under each parameter setting, in order to avoid randomness, we randomly generate 10 networks and report their average performance in subsequent experiments. 

\begin{table}[ht]
\begin{center}
    \caption{Statistics of the generated LFR artificial networks.}
	\label{tab:LFR}
	\begin{tabular}{cccccccc}
		\toprule
		$N$    & $\mu$   & $\langle k \rangle$ & $k_{max}$ & $C_{min}$ & $C_{max}$ & $\gamma$ & $\beta$  \\
		\midrule
		1,000 & 0.6 & 20 & 50 & 20 & 100 &2 & 1  \\
		1,000 & 0.7 & 20 & 50 & 20 & 100 &2 & 1  \\
		1,000 & 0.8 & 20 & 50 & 10 & 50  &2 & 1  \\
		\bottomrule
	\end{tabular}
\end{center}
\end{table}

For constructing synthetic attributed networks, we generate a $m$-dimensional binary attribute vector for each node according to its ground-truth label generated by the LFR benchmark, where the same $D (D < m)$ attributes are assigned to nodes in the same cluster. 
In this way, nodes can be perfectly grouped into ideal clusters in terms of attributes. In our experiments, we set $m = 100$ and $D = 10$ for testing high dimensional attributes. 
On the other hand, in order to verify the importance of node attributes, we add noise to blur the attribute cluster structure, which can be mimicked by randomly flipping a portion of attributes of each node (we call it noise ratio). With the increase of the noise ratio, the clearness of cluster structure decreases. Here, we set the noise ratio from 10\% to 40\% with a step length of 10\%.
These artificial attributed networks can clearly show the clearness of the cluster structures of network topology and node attributes. The experimental results on these networks are conducive to explaining the performance of different models and are able to avoid the influence of human bias on labelling real-world data.

\subsubsection{Evaluation metrics}

The clustering performance is evaluated by four popular metrics \cite{metrics}: Accuracy (ACC), Normalized Mutual Information (NMI), Average Rand Index (ARI), and F1-score (F1). As for these metrics, the larger value indicates the better clustering results.
They are defined as follows.

\textbf{Accuracy (ACC)} measures the correctly partitioned clusters $C$ according to the ground-truth ones $C^*$.
\begin{equation}
    ACC(C, C^*) = \frac{\sum_{i=1}^{\vert C \vert} \delta(C_i, C_i^*)}{\vert C \vert}
\end{equation}
where $\delta(a,b)$ denotes the Kronecker function which equals to one if $a = b$ and zero otherwise.

\textbf{Normalized Mutual Information (NMI)} quantifies the similarity between the ground-truth clusters $C^*$ and the inferred ones $C$.
\begin{equation}
    NMI(C, C^*) = \frac{2MI(C, C^*)}{H(C) + H(C^*)}
\end{equation}
where $MI(C, C^*)$ measures the mutual information between clusters $C$ and $C^*$, while $H(C)$ and $H(C^*)$ are the entropy of these two clusters, respectively.

\textbf{Average Rand Index (ARI)} is the enhancement clustering index of the rand index (RI), it is defined as follows.
\begin{equation}
    ARI = \frac{\sum_{ij}\binom{n_{ij}}{2} - [\sum_i\binom{n_{i \cdot}}{2} \sum_j\binom{n_{\cdot j}}{2}] / \binom{N}{2} }{\frac{1}{2}[\sum_i\binom{n_{i \cdot}}{2} + \sum_j\binom{n_{\cdot j}}{2}] - [\sum_i\binom{n_{i \cdot}}{2} \sum_j\binom{n_{\cdot j}}{2}] / \binom{N}{2}}
\end{equation}
where $N$ is the number of nodes, $n_{ij}$ represents the number of nodes belonging to both cluster $i$ and cluster $j$, $n_{i \cdot}$ and $n_{\cdot j}$ are the numbers of nodes in cluster $i$ and cluster $j$, respectively. 

\textbf{F1-score (F1)} is the harmonic mean of Precision and Recall. Let $T$ denotes the set of nodes in the ground-truth clusters and $S$ denotes the set of nodes assigned by a given algorithm in the corresponding clusters. $\vert \cdot \vert$ denotes the cardinality of a set. Then F1 can be formulated as below.
\begin{equation}
    F1 = \frac{2 \times precision \times recall}{precision + recall}
\end{equation}
where $precision = \vert S \cap T \vert / \vert S \vert$, $recall = \vert S \cap T \vert / \vert T \vert$.

\subsubsection{Baseline methods}

We compare our model with several state-of-the-art network embedding methods and deep embedded clustering models for clustering benchmark attributed networks to give a comprehensive study. 
For network embedding approaches, we run $K$-means to get the final clustering results on the learned representation vectors. All the compared models are listed below.
\begin{itemize}
    \item {\textbf{AE}}: It is the basis of our AE-based module, it reconstructs the node attributes to get node representations.
    \item {\textbf{GAE}} \cite{GAE}: It employs GCN as the encoder and reconstructs the link structure using the inner product decoder, which is the basis of our GAE-based module.
    \item {\textbf{DFCN}} \cite{DFCN}: It fuses the representations learned from two different views and combines triplet self-supervised clustering loss with reconstruction loss, guiding the whole learning process.
	\item {\textbf{DEC}} \cite{DEC}: It uses a stacked autoencoder to learn node representations using only node attributes, and then defines a clustering loss on the learned vectors to cluster the nodes. 
	\item {\textbf{IDEC}} \cite{IDEC}: It adds the reconstruction loss on the basis of the clustering loss of DEC, and optimizes the unified objective function jointly.
	\item {\textbf{DAEGC}} \cite{DAEGC}: It utilizes a graph attention network to capture the importance of nodes and carries out deep embedded clustering for attributed networks.
	\item {\textbf{SDCN}} \cite{SDCN}: It introduces a delivery operator to combine the GCN module for attributed graphs and the AE module for node attributes, then jointly optimizes node representations and clusters using a dual self-supervised module.
\end{itemize}

\subsubsection{Implementation details}

For the newly proposed DCP-DEC model, in order to jointly optimize the cluster assignments and node representations, we first pre-train AE and GAE models 50 epochs without the clustering part. That is to say, we simply minimize the reconstruction losses of Eqs.~\ref{l_are} and \ref{l_gre} in AE and GAE modules, respectively, as shown in step 1 of Algorithm~\ref{alg:algorithm}. Consequently, we obtain the initial parameters and the initial representations of two parts. Subsequently, the $K$-means algorithm is then performed on two groups of generated representation vectors 20 times, respectively. For each representation vector $\mathbf{Z}_a$ and $\mathbf{Z}_g$, we minimize the standard $K$-means error, and then select the cluster centers $\mu^a$ and $\mu^g$ corresponding to the best clustering result in 20 runs for initialization. 
After that, the entire model is jointly trained with three parts for representation learning and clustering. 
We set the final dimension of representation vectors in two DEC modules as 64. 
As for the encoder of GAE-based module, we set the hidden layer dimension to 512 for ACM and DBLP, and 256 for other real-world datasets and artificial attributed networks. 
Furthermore, we tune the number of latent layers and the dimension of each layer for AE-based module on real datasets. Consequently, each real dataset has different settings for the AE-based module. 
For artificial networks, in order to fully extract attribute representations of nodes while keeping the simplicity of our model, we set the dimension of hidden layers to 256-64 or 512-256-64, which respectively correspond to the networks with clear attribute clusters (noise ratio is 10\%) or unclear attribute clusters (noise ratio is 20\%, 30\%, or 40\%).
Besides, we use the grid search strategy to select the optimal combination of balance coefficients $\alpha, \beta, \gamma$, where each value ranges from 0 to 1. 
We use the Adam optimizer to train the whole model with the learning rate $10^{-3}$. 

For a fair comparison, with the exception of DAEGC, we replicate the other methods according to the optimal settings reported in the original paper.
Since DAEGC has no open source code, we directly report the results listed in the original paper for Citeseer, Cora, and PubMed networks, while for other real-world datasets, we use the results shown in SDCN.
We run our model and all other methods 10 times and report the average results to avoid extreme cases.

\subsection{Results and analysis}

\subsubsection{Performance of real-world datasets}

In this section, we validate the clustering performance of our proposed model compared with other state-of-the-art methods listed above on eight real-world datasets. 
We also report the performance of the other three DCP-DEC variants. Instead of using assignments $Q_g$ to obtain final clusters, we utilize DCP\_$Q_a$ to represent the clustering results generated through cluster distribution $Q_a$. What's more, we carry out K-means on two representations $\mathbf{Z}_g^{(L)}$ and $\mathbf{Z}_a^{(L)}$ learned from GAE and AE modules to obtain the other two groups of results, denoted as $Z_g$\_clu and $Z_a$\_clu, respectively.
All the results are shown in Table~\ref{tab:results}. Except for the three DCP-DEC variant results represented in the last three columns, the best results are highlighted in bold, and the sub-optimal ones are underlined. For the four versions of our DCP-DEC model, the best results are marked in italics.

\begin{table*}[ht]
	\scriptsize
	\begin{center}
		\caption{Comparison results on eight real-world datasets for clustering.}
		\label{tab:results}
		\resizebox{\linewidth}{!}{\begin{tabular}{c|c|cccccccc|ccc}
				\hline
				Dataset                   & Metric & AE              & GAE    & DEC    & IDEC         & DAEGC*          & SDCN         & DFCN            & DCP-DEC                 & DCP\_$Q_a$             & $Z_g$\_clu            & $Z_a$\_clu             \\
				\hline
				\multirow{4}{*}{Citeseer} & ACC    & 0.5665          & 0.4048 & 0.6048 & 0.5994       & 0.6720          & 0.6551       & \underline{0.6959}    & \textbf{0.7091}          & \textit{0.7105} & 0.7099 & \textit{0.7105} \\
				& NMI    & 0.2952          & 0.1318 & 0.3325 & 0.3271       & 0.3970          & 0.3814       & \underline{0.4382}    & \textbf{0.4487}          & \textit{0.4516} & 0.4508 & \textit{0.4516} \\
				& ARI    & 0.2816          & 0.1281 & 0.3390 & 0.3329       & 0.4100          & 0.3939       & \underline{0.4540}    & \textbf{0.4723}          & 0.4747 & 0.4742 & \textit{0.4748} \\
				& F1     & 0.5329          & 0.3742 & 0.5698 & 0.5662       & 0.6360          & 0.5986       & \underline{0.6426}    & \textbf{0.6611}          & \textit{0.6683} & 0.6659 & \textit{0.6683} \\
				\hline
				\multirow{4}{*}{Cora}     & ACC    & 0.4896          & 0.5211 & 0.4949 & 0.5099       & \textbf{0.7040} & 0.5495       & 0.6502          & \underline{\textit{0.6682}}    & 0.5722               & 0.6360               & 0.4726               \\
				& NMI    & 0.2724          & 0.3081 & 0.2639 & 0.3112       & \textbf{0.5280} & 0.3590       & \underline{0.5186}    & \textit{0.4709}          & 0.3349               & 0.4562               & 0.2992               \\
				& ARI    & 0.2116          & 0.2478 & 0.2230 & 0.2345       & \textbf{0.4960} & 0.2979       & 0.4289          & \underline{\textit{0.4375}}    & 0.2972               & 0.3976               & 0.2191               \\
				& F1     & 0.4919          & 0.4930 & 0.4679 & 0.4627       & \textbf{0.6820} & 0.4653       & \underline{0.6208}    & \textit{0.5901}          & 0.5171               & 0.5667               & 0.4724               \\
				\hline
				\multirow{4}{*}{PubMed}   & ACC    & 0.6192          & 0.6220 & 0.6063 & 0.6142       & \textbf{0.6710} & 0.6248       & 0.4792          & \underline{0.6426}             & 0.6327               & \textit{0.6697}      & 0.4066               \\
				& NMI    & \textbf{0.2674} & 0.2441 & 0.1984 & 0.2098       & \underline{0.2660}    & 0.2231       & 0.0727          & 0.2559                   & 0.2422               & \textit{0.2949}      & 0.0606               \\
				& ARI    & 0.2353          & 0.2341 & 0.1830 & 0.1953       & \textbf{0.2780} & 0.2104       & 0.0697          & \underline{0.2390}             & 0.2263               & \textit{0.2846}      & 0.0265               \\
				& F1     & 0.6069          & 0.6094 & 0.6076 & 0.6161       & \textbf{0.6590} & 0.6337       & 0.4233          & \underline{0.6375}             & 0.6287               & \textit{0.6645}      & 0.3620               \\
				\hline
				\multirow{4}{*}{ACM}      & ACC    & 0.8768          & 0.5407 & 0.8385 & 0.8401       & 0.8694          & 0.8842       & \underline{0.9078}    & \textbf{0.9138}          & 0.9173               & 0.9149               & \textit{0.9231}      \\
				& NMI    & 0.5957          & 0.2371 & 0.5018 & 0.5099       & 0.5618          & 0.6421       & \underline{0.6939}    & \textbf{0.7058}          & 0.7167               & 0.7075               & \textit{0.7267}      \\
				& ARI    & 0.6672          & 0.2497 & 0.5750 & 0.5787       & 0.5935          & 0.6925       & \underline{0.7480}    & \textbf{0.7618}          & 0.7709               & 0.7645               & \textit{0.7852}      \\
				& F1     & 0.8765          & 0.5110 & 0.8387 & 0.8407       & 0.8707          & 0.8836       & \underline{0.9072}    & \textbf{0.9138}          & 0.9170               & 0.9150               & \textit{0.9232}      \\
				\hline
				\multirow{4}{*}{DBLP}     & ACC    & 0.5366          & 0.4727 & 0.6025 & 0.6091       & 0.6205          & 0.6682       & \underline{0.7644}    & \textbf{0.7751}          & \textit{0.7755}      & 0.7686               & 0.6241               \\
				& NMI    & 0.2296          & 0.1427 & 0.2533 & 0.2591       & 0.3249          & 0.3215       & \underline{0.4434}    & \textit{\textbf{0.4699}} & 0.4691               & 0.4653               & 0.3414               \\
				& ARI    & 0.1348          & 0.1192 & 0.2491 & 0.2545       & 0.2103          & 0.3406       & \underline{0.4765}    & \textbf{0.4904}          & \textit{0.4912}      & 0.4781               & 0.2162               \\
				& F1     & 0.5049          & 0.4675 & 0.6006 & 0.6067       & 0.6175          & 0.6559       & \underline{0.7616}    & \textbf{0.7738}          & \textit{0.7745}      & 0.7678               & 0.6237               \\
				\hline
				\multirow{4}{*}{USPS}     & ACC    & 0.6888          & 0.6347 & 0.7451 & 0.7484       & 0.7355          & 0.7800       & \underline{0.7919}    & \textbf{0.8881}          & 0.8828               & \textit{0.8920}      & 0.8824               \\
				& NMI    & 0.6390          & 0.6245 & 0.7362 & 0.7624       & 0.7112          & 0.7899       & \textbf{0.8197} & \underline{0.8108}             & 0.8032               & \textit{0.8193}               & 0.8040               \\
				& ARI    & 0.5436          & 0.5243 & 0.6539 & 0.6756       & 0.6333          & 0.7104       & \underline{0.7450}    & \textbf{0.7846}          & 0.7802               & \textit{0.7929}      & 0.7797               \\
				& F1     & 0.6819          & 0.6113 & 0.7373 & 0.7419       & 0.7245          & 0.7663       & \underline{0.7772}    & \textbf{0.8860}          & 0.8789               & \textit{0.8900}      & 0.8785             \\
				\hline
				\multirow{4}{*}{HHAR}     & ACC    & 0.6736          & 0.6902 & 0.7110 & 0.7150       & 0.7651          & 0.8348       & \textbf{0.8700} & \underline{0.8384}             & 0.8434               & 0.8390               & \textit{0.8436}      \\
				& NMI    & 0.6734          & 0.6678 & 0.6715 & 0.7292       & 0.6910          & \underline{0.7988} & \textbf{0.8217} & 0.7754                   & 0.7684               & \textit{0.7765}      & 0.7682               \\
				& ARI    & 0.5822          & 0.5580 & 0.5712 & 0.6046       & 0.6038          & \underline{0.7278} & \textbf{0.7625} & 0.7068                   & 0.7038               & \textit{0.7099}      & 0.7041               \\
				& F1     & 0.6600          & 0.6805 & 0.7073 & 0.6940       & 0.7689          & 0.8130       & \textbf{0.8728} & \underline{0.8316}             & 0.8427               & 0.8339               & \textit{0.8429}      \\
				\hline
				\multirow{4}{*}{Reuters}  & ACC    & 0.7551          & 0.4174 & 0.7316 & 0.7588       & 0.6550          & 0.7542       & \underline{0.7774}    & \textbf{0.8033}          & 0.7657               & \textit{0.8048}      & 0.7667               \\
				& NMI    & 0.4795          & 0.1288 & 0.4604 & 0.4938       & 0.3055          & 0.4784       & \textbf{0.6019} & \underline{0.5175}             & 0.5314               & \textit{0.5246}      & 0.5317               \\
				& ARI    & 0.5151          & 0.0542 & 0.4800 & 0.5451       & 0.3112          & 0.5110       & \underline{0.5985}    & \textbf{0.6064}          & 0.5584               & \textit{0.6097}      & 0.5619               \\
				& F1     & 0.6860          & 0.3100 & 0.6765 & \underline{0.6965} & 0.6182          & 0.6348       & 0.6964          & \textbf{0.7107}          & 0.6905               & \textit{0.7133}      & 0.6911           \\
				\hline 
		\end{tabular}}
	\end{center}
\end{table*}

From Table~\ref{tab:results}, we can summarize that our proposed model achieves superior or competitive performance on most benchmarks across the evaluation metrics. 
More specifically, we have the following observations.
\begin{itemize}
	\item The one-stage DEC models consistently outperform the compared network embedding methods for clustering. Although DFCN \cite{DFCN} looks like a two-stage method, it employs a triplet self-supervised strategy as the guidance for node clustering during the process of representation learning, which uses the same idea as one-stage end-to-end models. This verifies that with the explicit clustering constraint, the learned representation vectors can produce better clustering results.
	\item For deep embedded clustering models, DEC \cite{DEC} and IDEC \cite{IDEC} only utilize node features, so their performance is worse than other methods that fuse topological and attribute information. This also proves the effectiveness of considering two kinds of information in attributed networks comprehensively.
	\item Compared with the recently proposed methods, our model shows significant improvements over SDCN \cite{SDCN}. This strongly verifies our two intuitions. 1) Reconstructing network topology is helpful. 2) The cluster distribution constraint is a good guidance to learn the representations of nodes and their cluster structure. Though our DCP-DEC model is inferior to DAEGC \cite{DAEGC} on Cora and PubMed networks, and worse than DFCN \cite{DFCN} on HHAR, the performance is still competitive. Moreover, the DCP-DEC model is simple compared to DFCN.
	\item As for the four variants of the DCP-DEC model, the results are satisfactory if we directly use the cluster assignments as the final clustering. In most cases, the cluster distributions of $\mathbf{Q}_g$ and that of $\mathbf{Q}_a$ tend to be consistent, which proves the effectiveness of the distribution consistency constraint. Meanwhile, the results of $\mathbf{Q}_g$ are stronger than $\mathbf{Q}_a$, indicating that it is reasonable to select $Q_g$ as a guiding target. In addition, if we perform K-means on the learned representations, the model achieves even better results on some datasets such as PubMed and USPS, which shows that during our joint optimization, representation learning and clustering can promote each other.
\end{itemize}

\subsubsection{Performance of artificial attributed networks}

Since the newly proposed SDCN \cite{SDCN} and DFCN \cite{DFCN} have better clustering results than other methods, in this section, we further compare them with our newly proposed model DCP-DEC on artificial attributed networks to avoid the human bias on labelling datasets and give a more fair comparison.
We report the clustering performance in Table~\ref{tab: artificial}, where we highlight the best results in bold, and LFR* represents attributed networks generated by the LFR benchmark with the noise ratio $*\%$.
As mentioned before, DFCN further conducts $K$-means on the obtained node representation vectors to achieve clustering. We use `Z' to represent these DFCN's results in Table~\ref{tab: artificial}. Meanwhile, `Q' indicates that the soft cluster assignments updated during training are directly used as the clustering results of DFCN, which corresponds to the joint training mechanism in DEC models. `DCP' represents our DCP-DEC model in Table~\ref{tab: artificial} for space limitation.

\begin{table*}[ht]
\begin{center}
\caption{Comparison of clustering results on artificial attributed networks.}
\label{tab: artificial}
\resizebox{\linewidth}{!}{\begin{tabular}{c|c|c|cc|c|c|cc|c|c|cc|c}
\hline
\multirow{3}{*}{Dataset}   & \multirow{3}{*}{Metric} & \multicolumn{4}{c|}{$\mu=0.6$}        & \multicolumn{4}{c|}{$\mu=0.7$}        & \multicolumn{4}{c}{$\mu=0.8$}      \\
\cline{3-14}
    &                         & \multirow{2}{*}{SDCN} & \multicolumn{2}{c|}{DFCN} & \multirow{2}{*}{DCP} & \multirow{2}{*}{SDCN} & \multicolumn{2}{c|}{DFCN} & \multirow{2}{*}{DCP} & \multirow{2}{*}{SDCN} & \multicolumn{2}{c|}{DFCN} & \multirow{2}{*}{DCP} \\
\cline{4-5}
\cline{8-9}
\cline{12-13}
    &             &          & Z               & Q      &                          &                       & Z               & Q      &                           &                       & Z               & Q      &         \\
\hline
\multirow{4}{*}{LFR10} & ACC     & 0.4874 & 0.9718          & 0.9538 & \textbf{0.9742} & 0.3929 & 0.9269          & 0.8896 & \textbf{0.9539} & 0.2635 & 0.6959          & 0.5532 & \textbf{0.8340} \\
    & NMI                     & 0.5866 & 0.9870           & 0.9668 & \textbf{0.9883} & 0.4749 & 0.9568          & 0.9044 & \textbf{0.9607} & 0.4114 & 0.7865          & 0.6780  & \textbf{0.8661} \\
    & ARI                     & 0.3420  & \textbf{0.9703} & 0.9361 & 0.9658          & 0.2325 & 0.9077          & 0.8347 & \textbf{0.9213} & 0.1147 & 0.6253          & 0.4442 & \textbf{0.6887} \\
    & F1                      & 0.3769 & \textbf{0.9662} & 0.9506 & 0.9606          & 0.3140  & 0.9174          & 0.8821 & \textbf{0.9415} & 0.1994 & 0.6210           & 0.4880  & \textbf{0.8206} \\
\hline
\multirow{4}{*}{LFR20} & ACC     & 0.4681 & 0.9646 & 0.9394 & \textbf{0.9840}         & 0.3823 & 0.9362   & 0.8921 & \textbf{0.9509}     & 0.2439 & 0.6299  & 0.4932 & \textbf{0.8523} \\
    & NMI                     & 0.5801 & 0.9859 & 0.9584 & \textbf{0.9921}          & 0.4541 & \textbf{0.9580}  & 0.9001 & 0.9414          & 0.3899 & 0.7288 & 0.6283 & \textbf{0.8764}  \\
    & ARI                     & 0.3268 & 0.9690  & 0.9286 & \textbf{0.9847}          & 0.2162 & \textbf{0.9218} & 0.8426 & 0.9052          & 0.1034 & 0.5392 & 0.3745 & \textbf{0.7588}  \\
    & F1                      & 0.3493 & 0.9521 & 0.9251 & \textbf{0.9640}          & 0.2853 & 0.9218 & 0.8772 & \textbf{0.9348}         & 0.1675 & 0.5530           & 0.4318 & \textbf{0.7633} \\
\hline
\multirow{4}{*}{LFR30} & ACC                     & 0.4636 & 0.9436          & 0.9000 & \textbf{0.9627} & 0.3632 & 0.8717          & 0.7613 & \textbf{0.8951} & 0.2042 & 0.4625          & 0.3315 & \textbf{0.6301} \\
    & NMI                     & 0.5811 & \textbf{0.9765} & 0.9313 & 0.9729          & 0.4337 & \textbf{0.9064} & 0.8085 & 0.8778          & 0.3307 & 0.5903          & 0.4924 & \textbf{0.6903} \\
    & ARI           & 0.3353 & 0.9508          & 0.8799 & \textbf{0.9540} & 0.2017 & \textbf{0.8581} & 0.7163 & 0.8236          & 0.0700   & 0.3373          & 0.1997 & \textbf{0.3661} \\
    & F1                      & 0.3294 & 0.9240  & 0.8608 & \textbf{0.9268}          & 0.2584 & 0.8339          & 0.7108 & \textbf{0.8346} & 0.136  & 0.3993          & 0.2893 & \textbf{0.5054} \\
\hline
\multirow{4}{*}{LFR40} & ACC                     & 0.4051 & 0.8966          & 0.7242 & \textbf{0.9199} & 0.2983 & 0.7294          & 0.5141 & \textbf{0.7678} & 0.1718 & 0.2746          & 0.2034 & \textbf{0.3037} \\
    & NMI                     & 0.4751 & \textbf{0.9394} & 0.8105 & 0.9224          & 0.3387 & 0.7559          & 0.5795 & \textbf{0.7756} & 0.2603 & \textbf{0.4308} & 0.3665 & 0.4128          \\
    & ARI           & 0.2383 & \textbf{0.8974} & 0.6935 & 0.8746          & 0.1333 & \textbf{0.6644} & 0.4179 & 0.6277          & 0.0393 & \textbf{0.1400}   & 0.0834 & 0.0929\\
    & F1            & 0.2640  & 0.8618          & 0.6619 & \textbf{0.9039} & 0.1862 & \textbf{0.6790}  & 0.4606 & 0.6167          & 0.1062 & \textbf{0.2433} & 0.1823 & 0.2260   \\
\hline
\end{tabular}}
\end{center}
\end{table*}

From Table~\ref{tab: artificial}, we can see that our proposed DCP-DEC model gets the best performance in most cases.
Note that the cluster structure of a network is unclear when $\mu=0.8$, and the number of clusters in an artificial network is much larger than that in real networks. This increases the difficulty of clustering. However, our model still achieves significantly better performance compared with DFCN or SDCN.

Specifically, though SDCN introduces a delivery operator to integrate the autoencoder-specific representation into structure-aware representation of GCN, from our results, it may not effectively combine network topology and node attributes, thus its results are much worse than those of DPC-DEC even when the cluster structure of an attributed network is clear (for example, LFR10 with $\mu=0.6$). 
In addition, DCP-DEC is comparable or even better than DFCN in most cases. 
We can draw the conclusion that our distribution consistency constraint between the two cluster assignments of DEC modules can well maintain the consistency of cluster structures obtained from topology and attributes. 
In other words, the cluster distribution consistency constraint is more robust or can be a substitute for fusing two representation vectors before clustering as DFCN does. 
However, once we use the soft cluster assignments of DFCN as final clustering results (the `Q' column in Table~\ref{tab: artificial}), the performance decreases significantly. This means that DFCN cannot directly well reveal the cluster structure of a given attributed network like DCP-DEC. All the results in synthetic networks further demonstrate the effectiveness of our DCP-DEC model for clustering tasks.

\begin{figure*}[ht]
	\centering
	\subfloat[Citeseer]{
		\includegraphics[width=3.73cm]{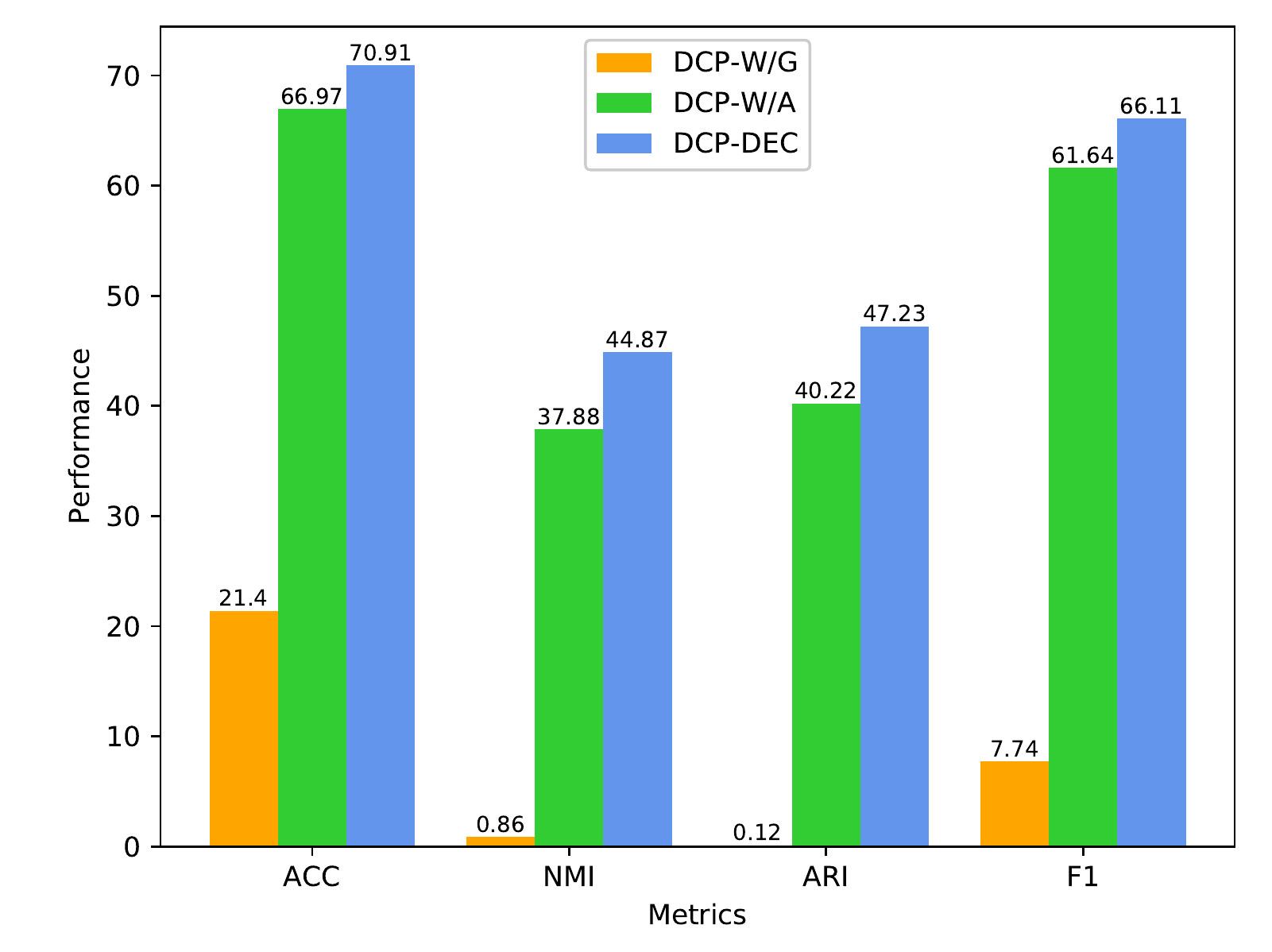}}
	\quad
	\subfloat[Cora]{
		\includegraphics[width=3.73cm]{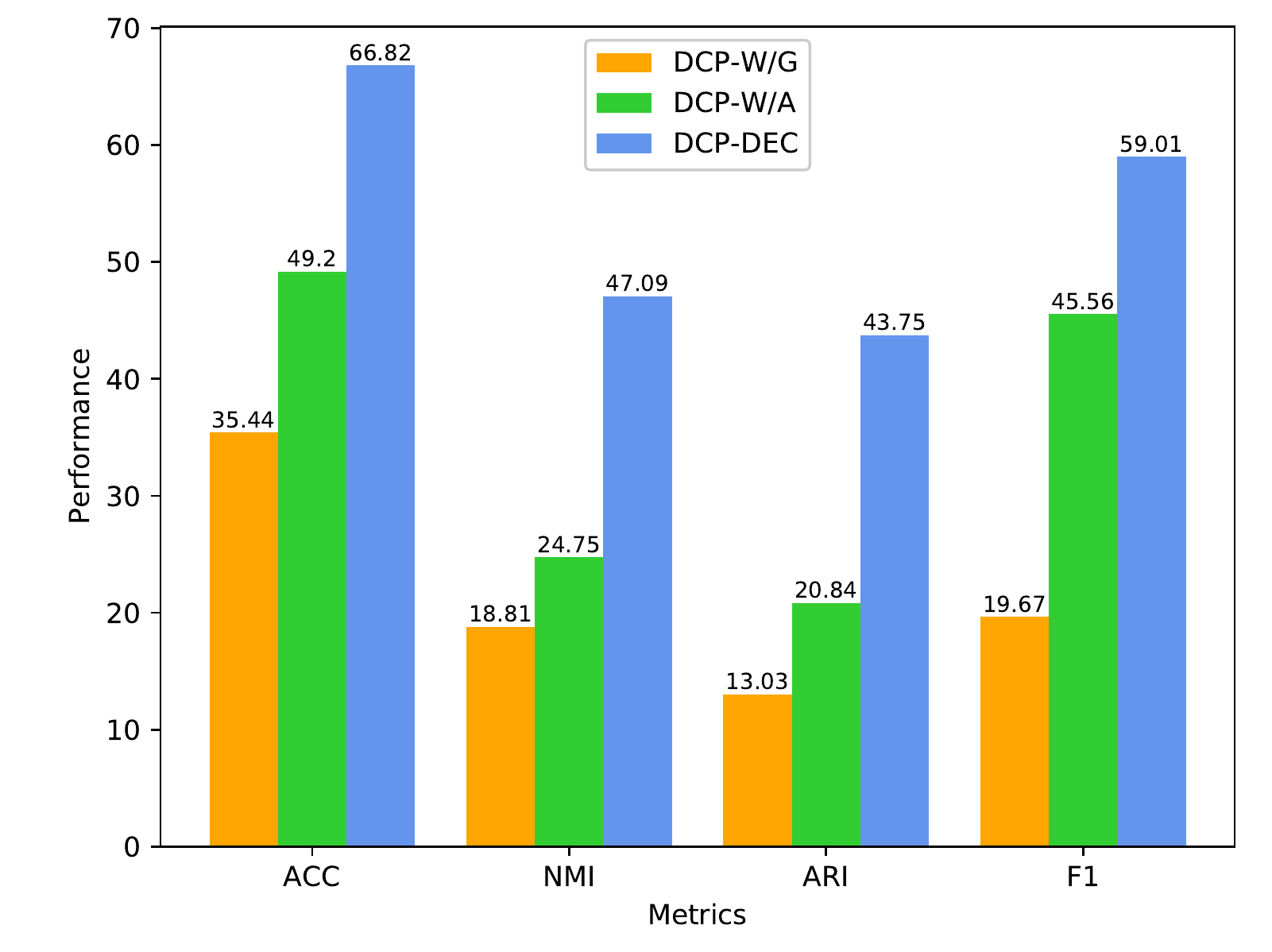}}
	\quad
	\subfloat[PubMed]{
		\includegraphics[width=3.73cm]{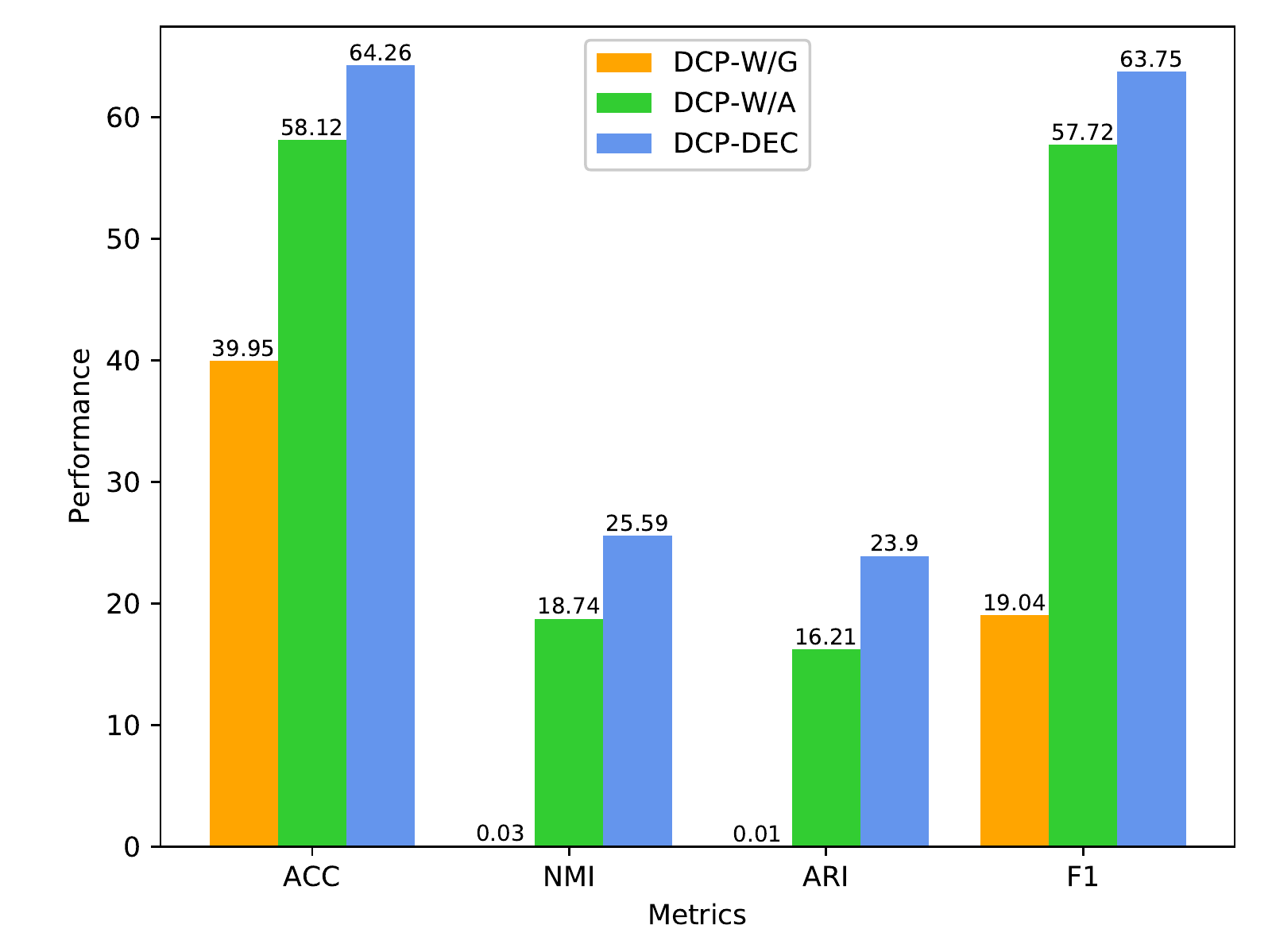}}
	\quad
	\subfloat[ACM]{
		\includegraphics[width=3.73cm]{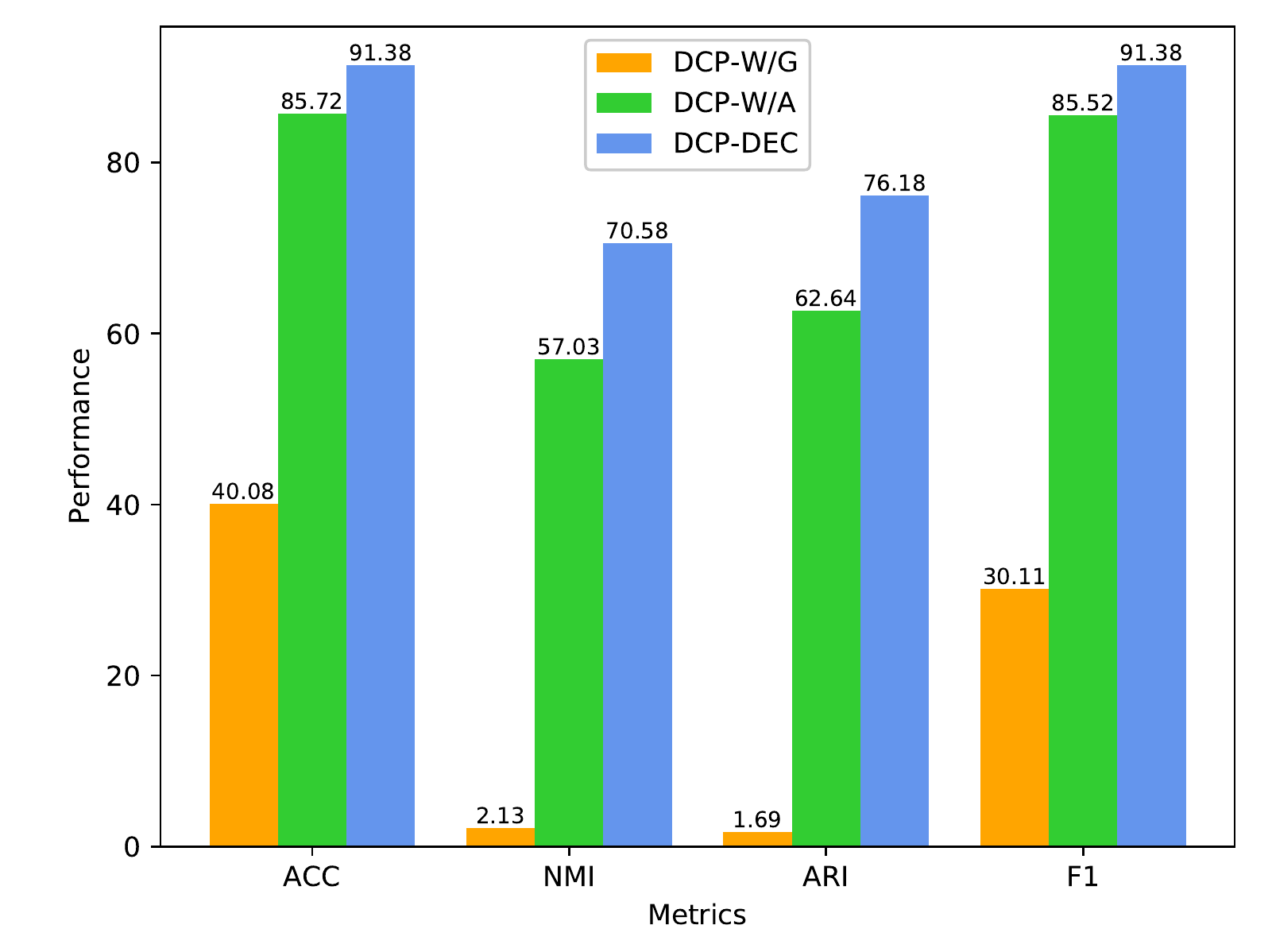}}
	\quad
	\subfloat[DBLP]{
		\includegraphics[width=3.73cm]{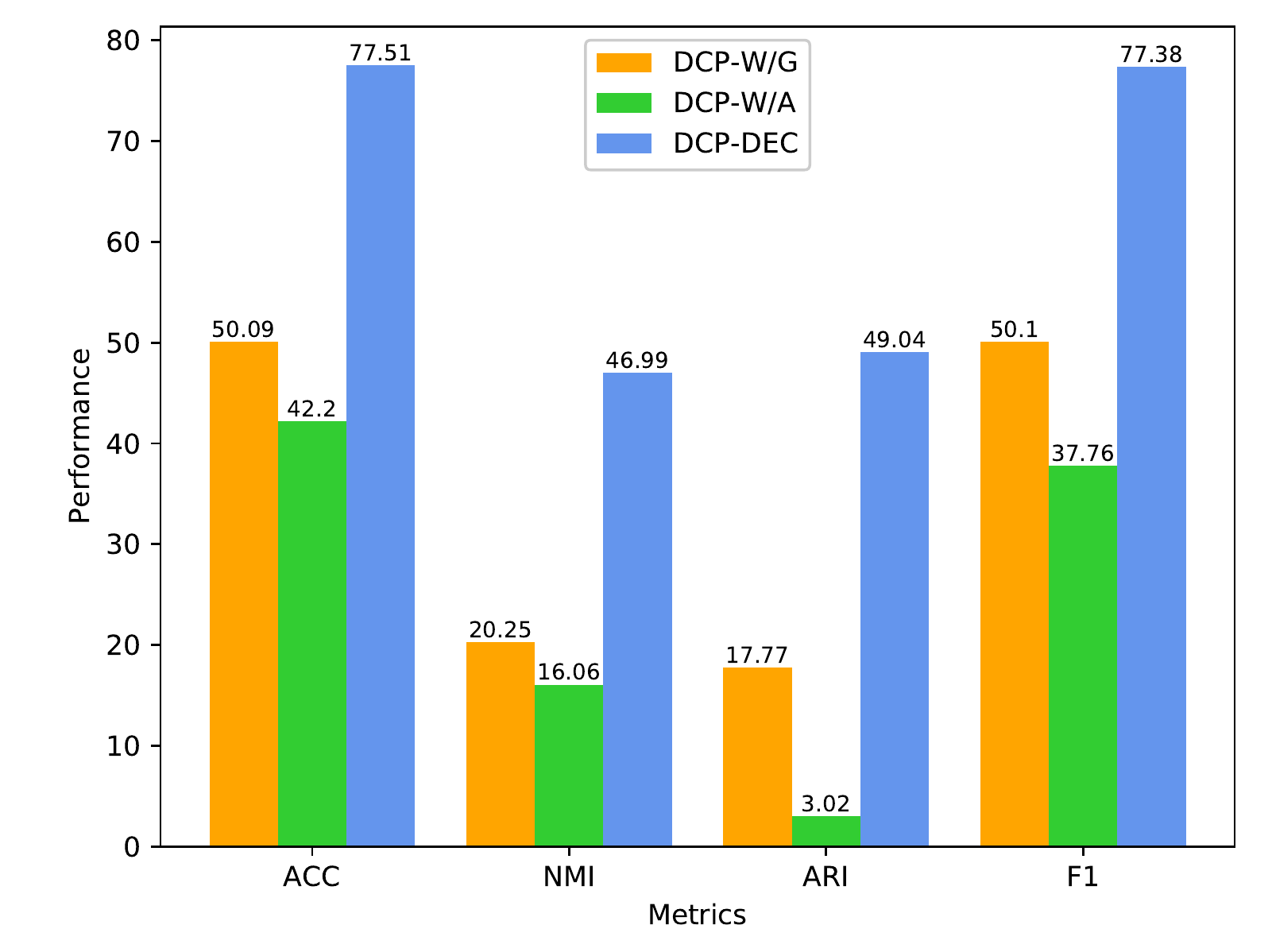}}
	\quad
	\subfloat[USPS]{
		\includegraphics[width=3.73cm]{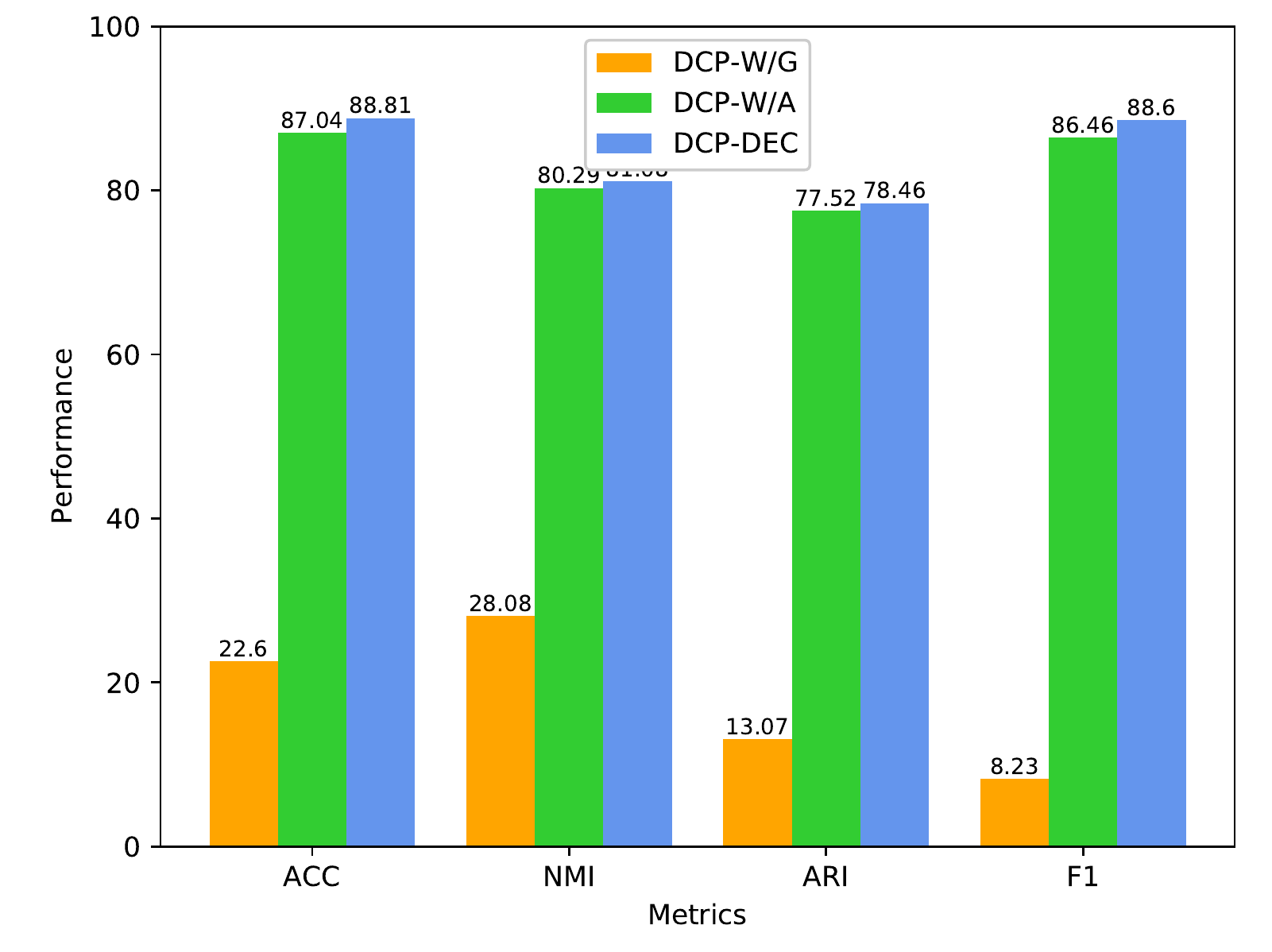}}
	\quad
	\subfloat[HHAR]{
		\includegraphics[width=3.73cm]{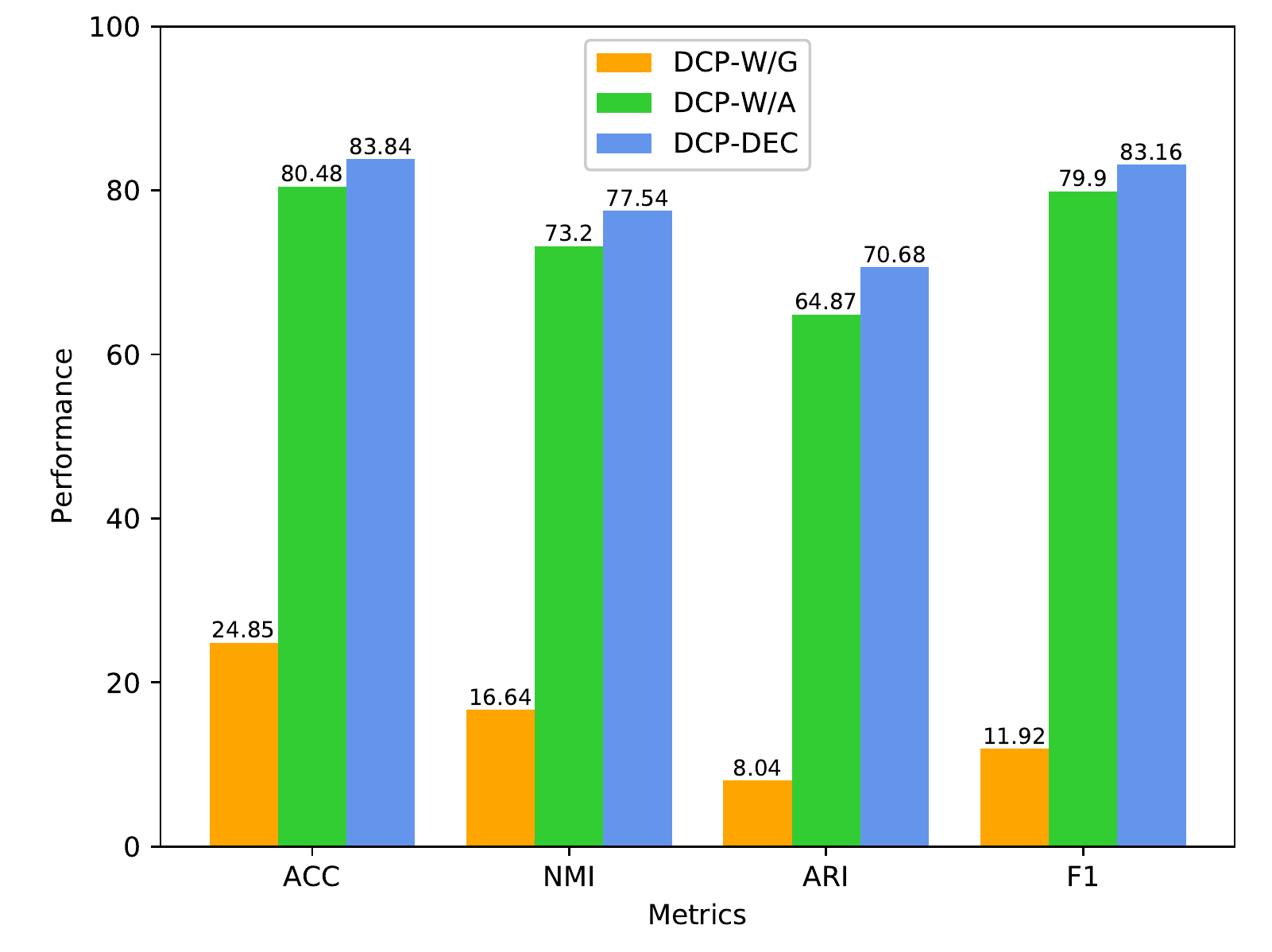}}
	\quad
	\subfloat[Reuters]{
		\includegraphics[width=3.73cm]{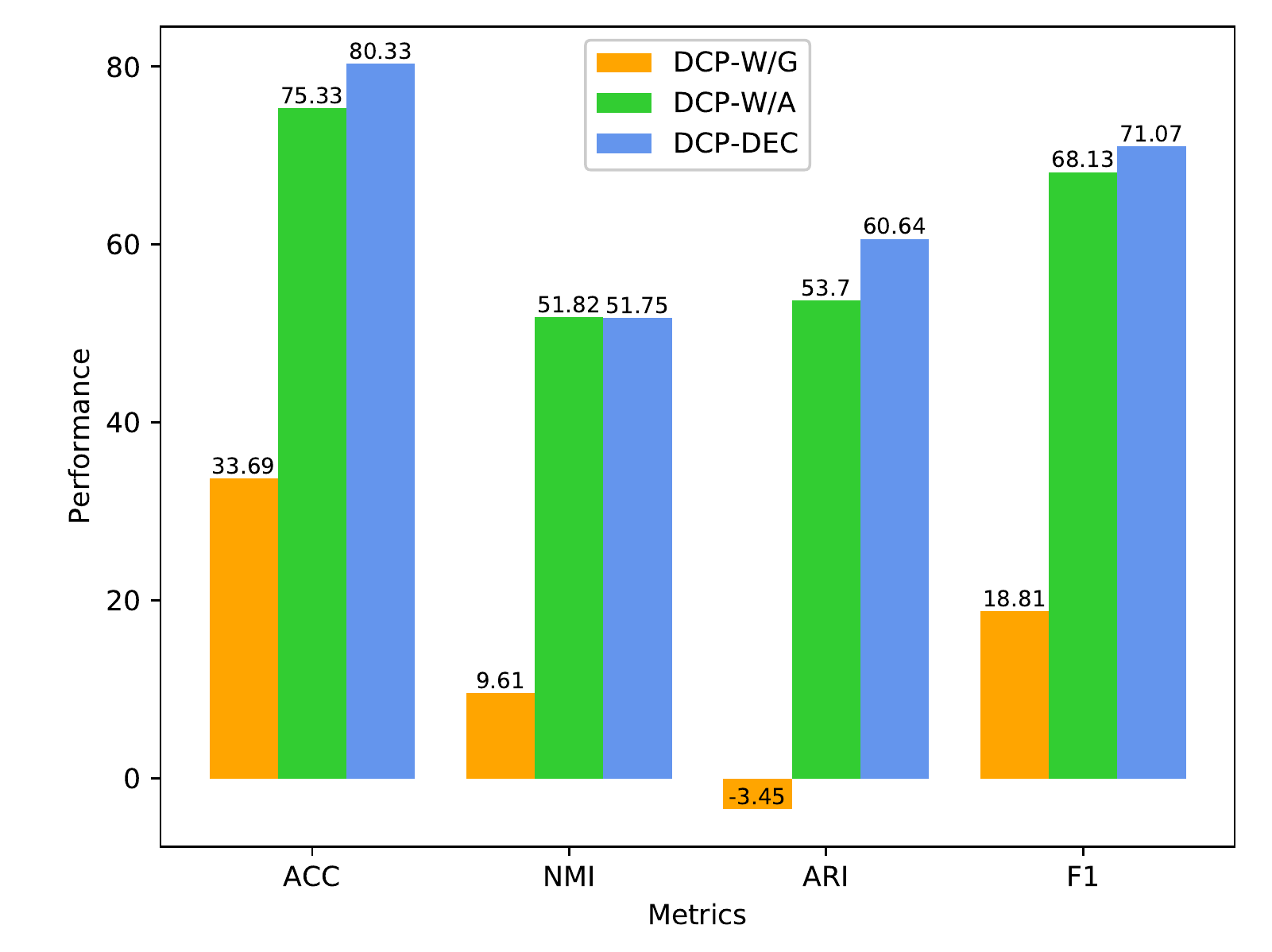}}
	\caption{Comparison results of DCP-DEC model with/without distribution consistency constraint.}
	\label{fig:ablation}
\end{figure*}

\subsection{Ablation study}

In this section, we aim to prove the effectiveness of the distribution consistency constraint in the whole DCP-DEC model. Once we remove this constraint (i.e., $\gamma = 0$), there are two cluster distributions from two different deep embedded clustering modules. Therefore, we compare these two clustering results with the final results of the whole DCP-DEC model (showed in
Fig.~\ref{fig:ablation}) on eight real-world datasets, where DCP-W/G and DCP-W/A respectively represent the clustering results of GAE-based module and AE-based module without the consistency constraint. 

According to Fig.~\ref{fig:ablation}, we have the following results. 1) The results of DCP-DEC model are significantly better than DCP-W/G and DCP-W/A. This indicates the effectiveness of the distribution consistency constraint used in DCP-DEC.
2) The results of DCP-W/A are usually better than DCP-W/G with the exception of DBLP network, verifying the importance of a deep autoencoder on node attributes. 
3) The overall performance of DCP-DEC demonstrates that our framework takes advantage of network topology and node attributes, and provides an effective way to combine two cluster assignments obtained from two views. All in all, the distribution consistency constraint plays an important role in our whole model.

\subsection{Visualization of training}
\begin{figure*}[ht]
	\centering
	\subfloat[ACM]{
		\includegraphics[width=5cm]{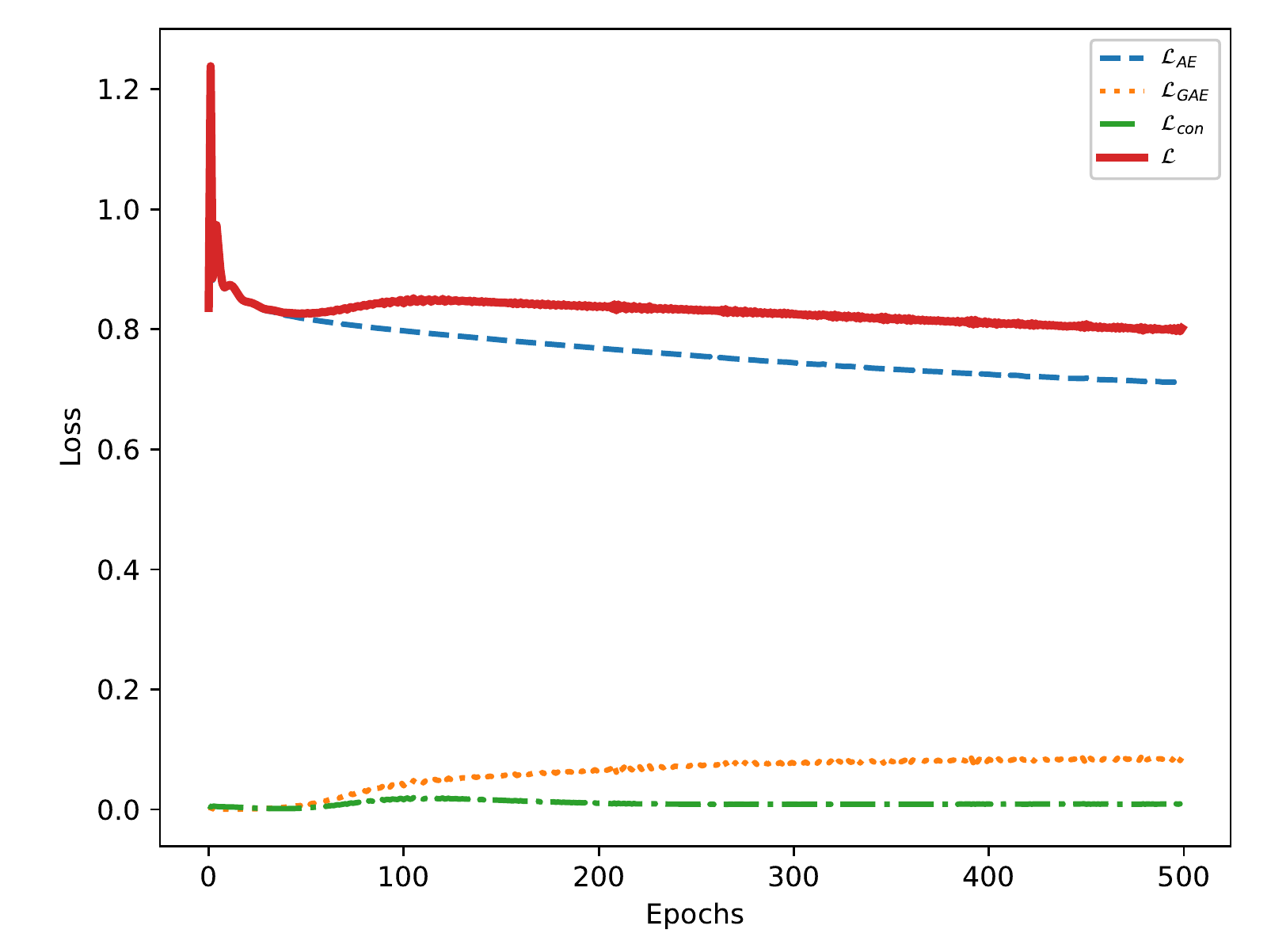}}
	\quad
	\subfloat[DBLP]{
		\includegraphics[width=5cm]{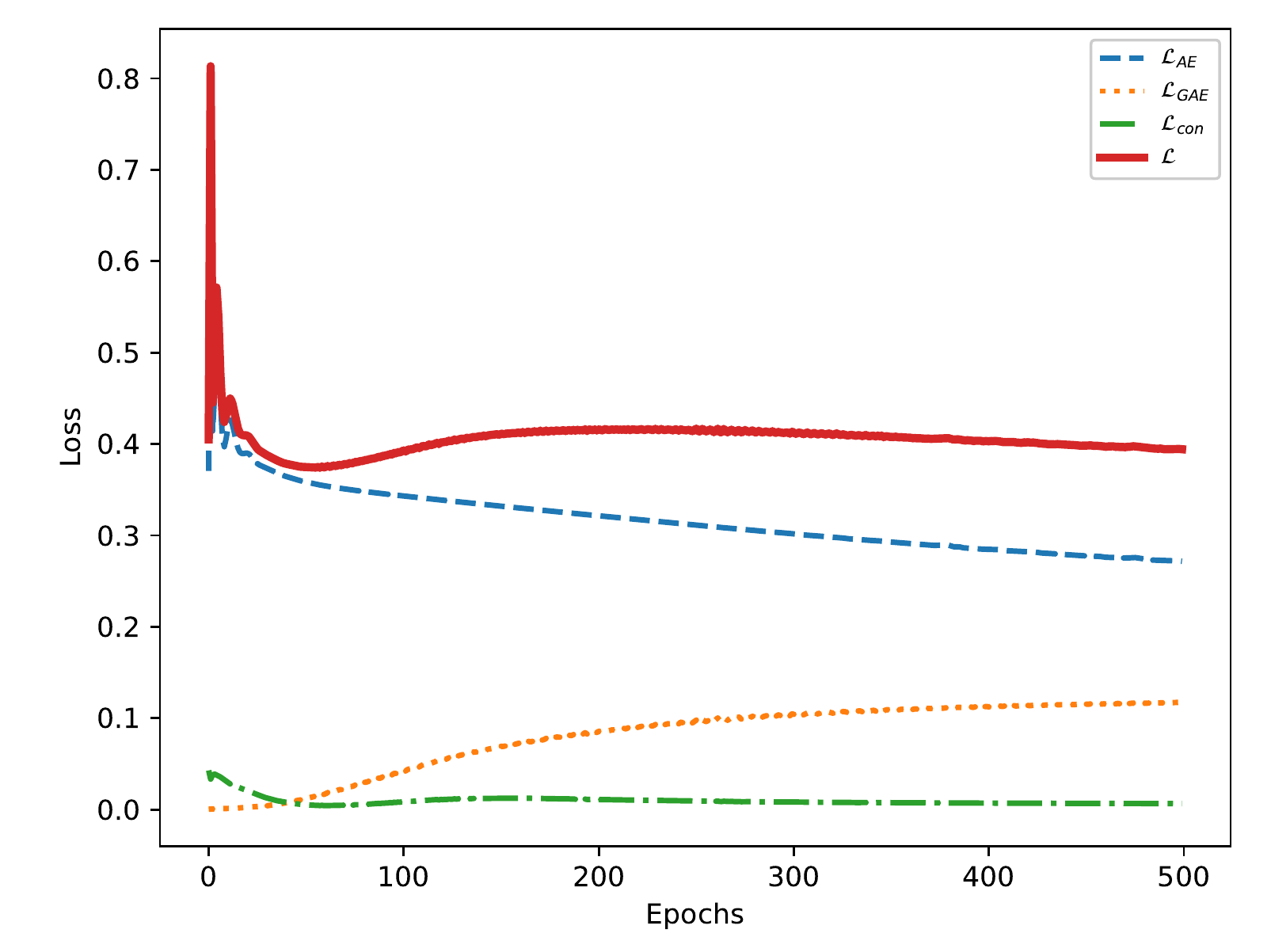}}
	\quad
	\subfloat[USPS]{
		\includegraphics[width=5cm]{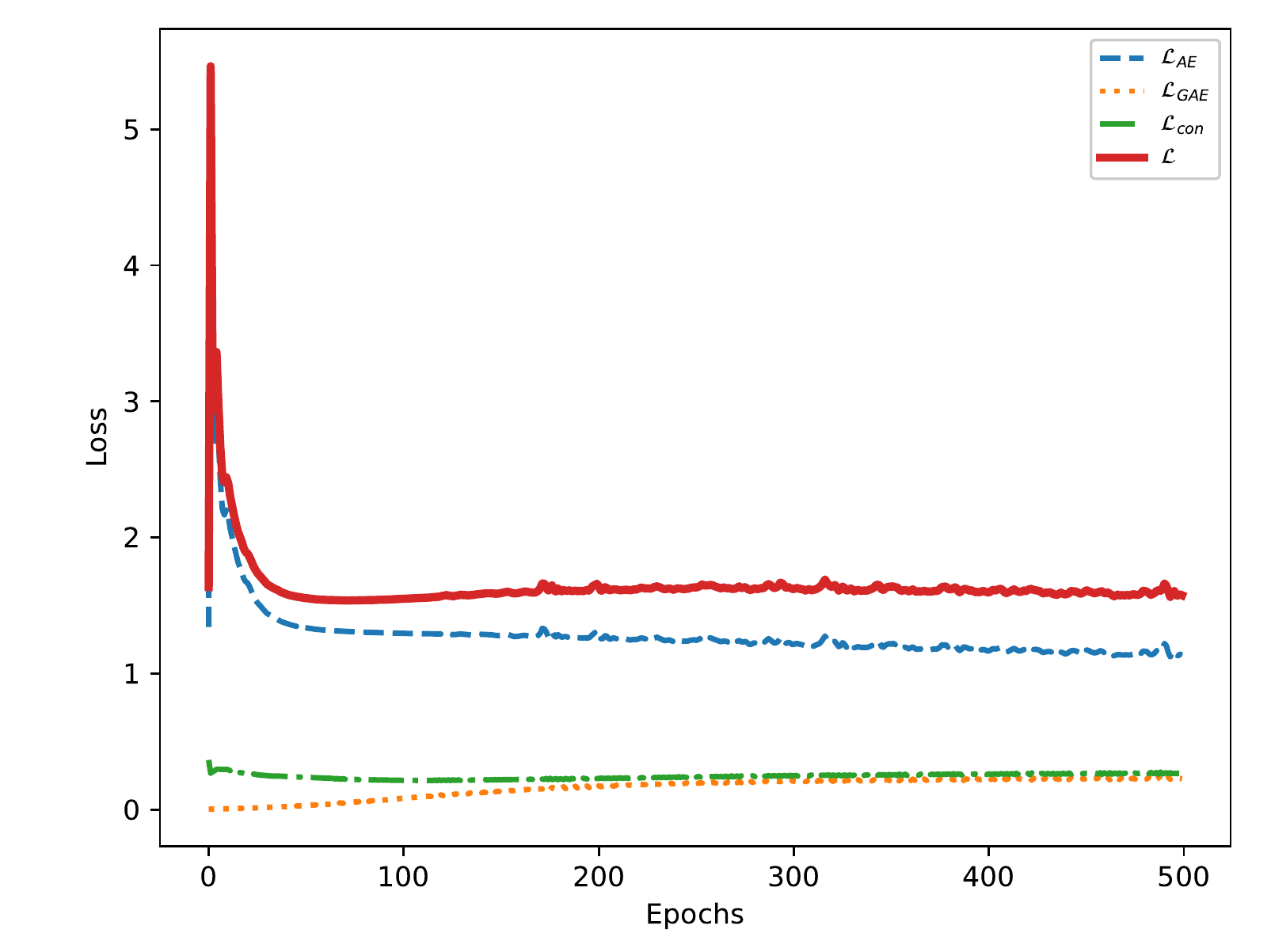}}
	\quad
	\subfloat[ACM]{
		\includegraphics[width=5cm]{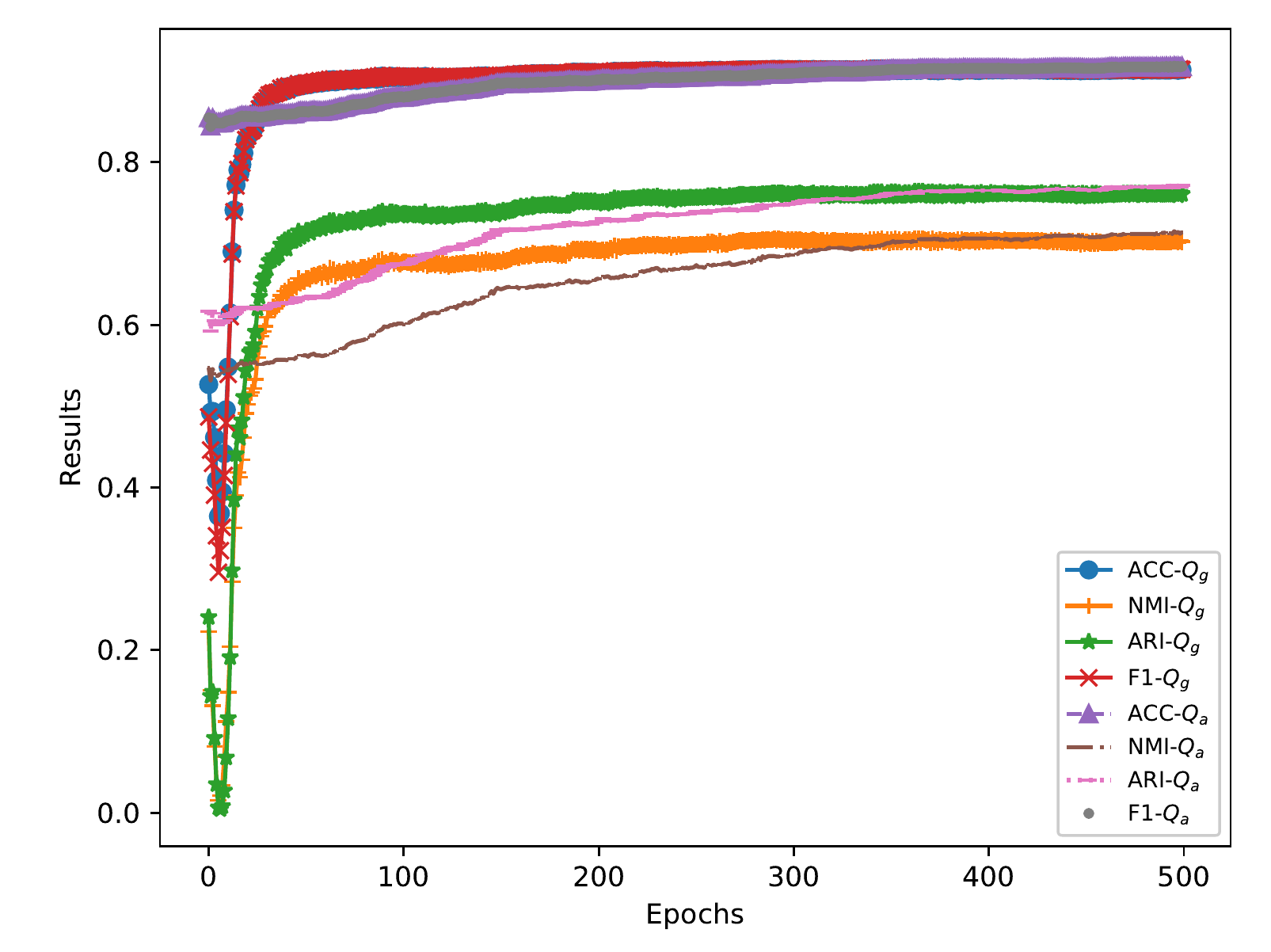}}
	\quad
	\subfloat[DBLP]{
		\includegraphics[width=5cm]{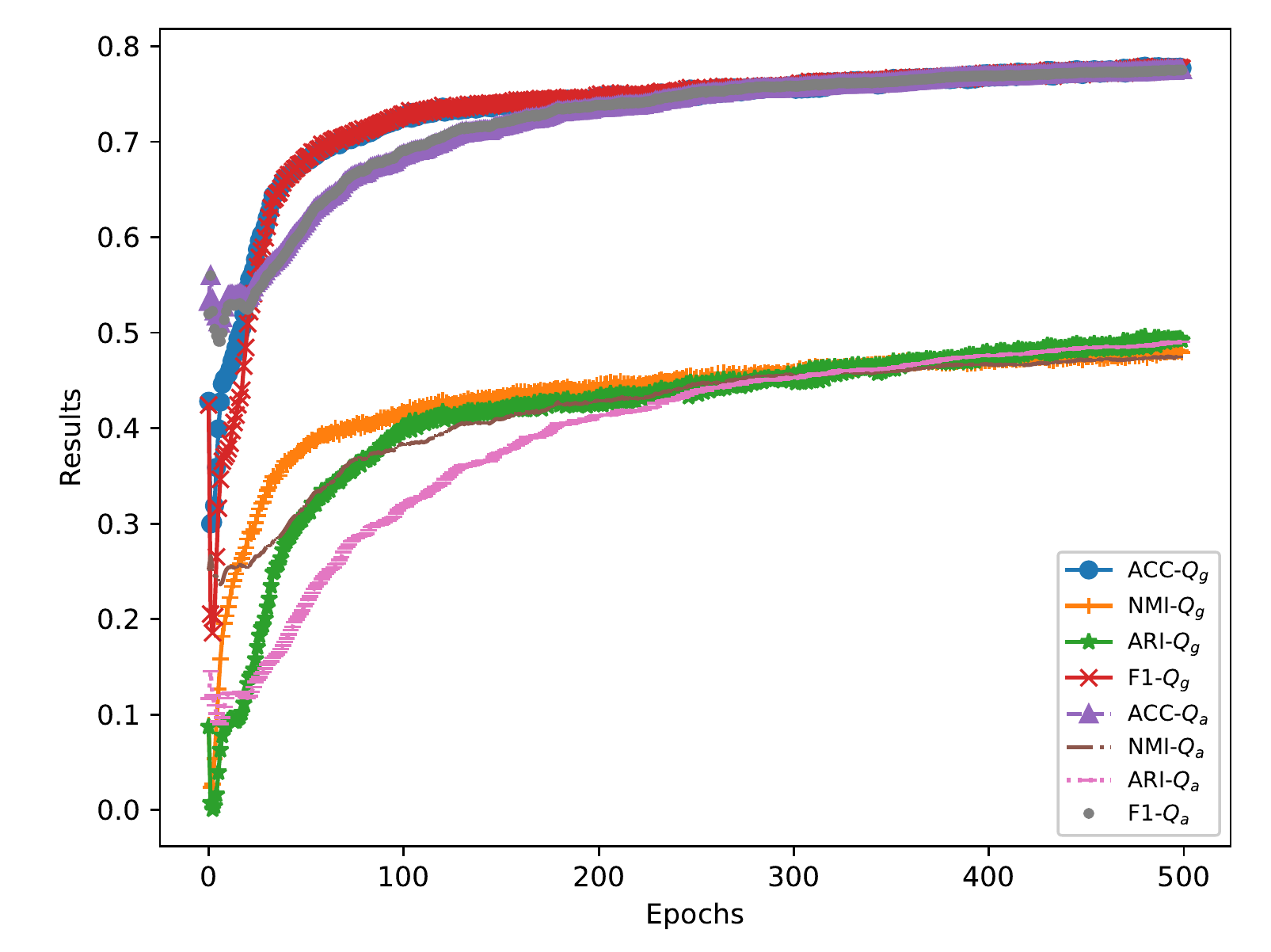}}
	\quad
	\subfloat[USPS]{
		\includegraphics[width=5cm]{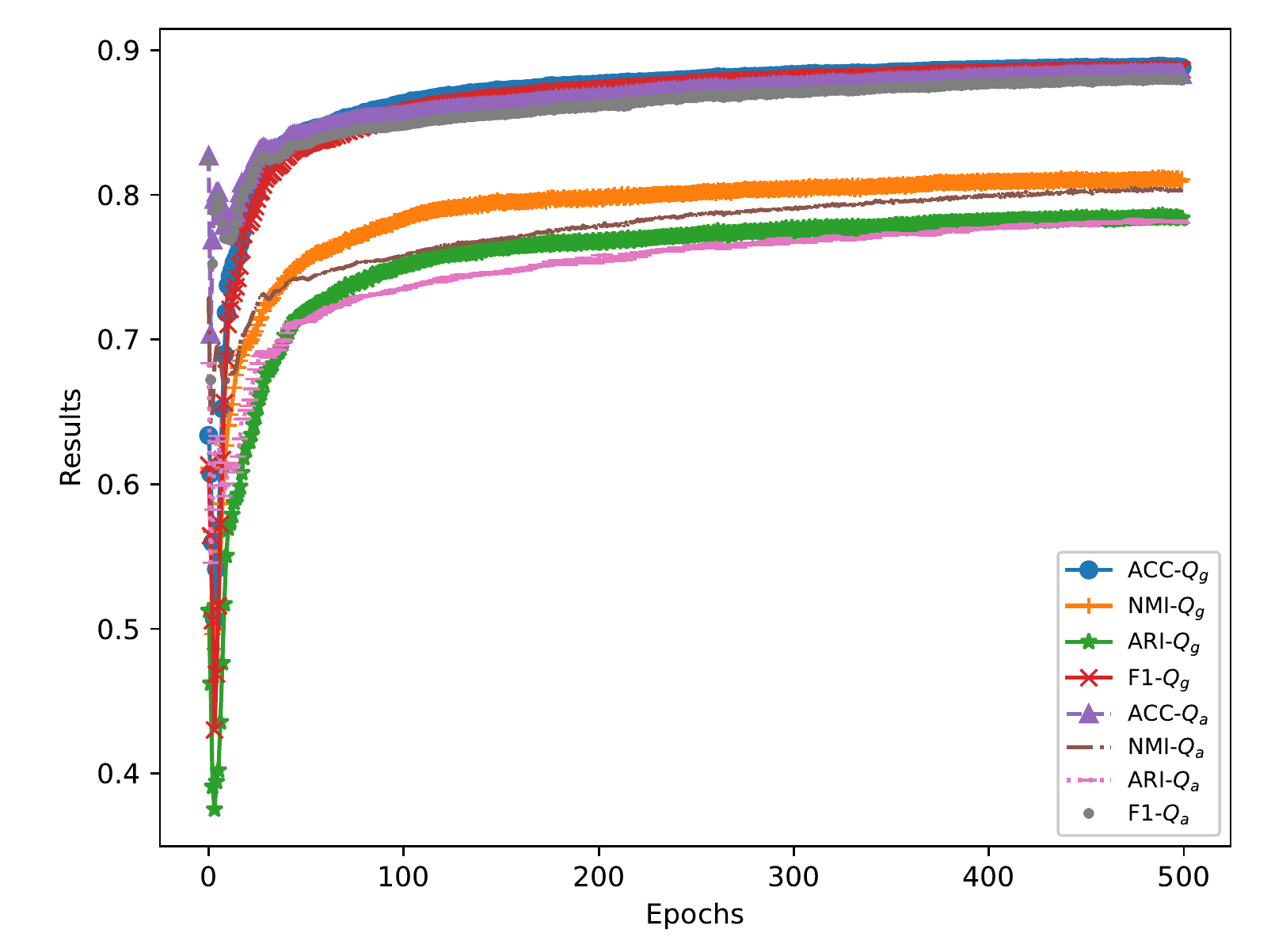}}
	\caption{Visualization of losses and evaluation results over training epochs.}
	\label{fig:loss-acc}
\end{figure*}

In order to have a more intuitive understanding of the training process and the convergence of the model, we visualize the loss of the whole objective function and that of each part of the function in Fig.~\ref{fig:loss-acc} (a)-(c). What’s more, the curves of the evaluation results for each measurement based on the cluster assignments $\mathbf{Q}_g$ and $\mathbf{Q}_a$ over the training epochs are also shown accordingly in Fig.~\ref{fig:loss-acc} (d)-(f). Since all the datasets have similar tendencies, we randomly take three real-world datasets, including ACM, DBLP, and USPS, as examples to illustrate and analyze the results.

From Fig.~\ref{fig:loss-acc}, we can see that in the first few epochs, the total loss value and accuracy change rapidly. However, after several rounds of iterations, the loss decreases steadily and can quickly reach convergence. At the same time, the performance of the DCP-DEC model gradually increases and reaches the optimum when the training converges. With the help of distribution consistency constraint, the cluster assignments of GAE and AE modules gradually tend to be consistent, illustrating the importance of end-to-end joint optimization in deep embedding clustering.

\subsection{Parameter sensitivity analysis}
To show the clustering performance of the proposed model, we also study the sensitivity of the parameters used in the experiments.

\subsubsection{Effect of the final dimension}

\begin{figure*}[ht]
    \centering
    \subfloat[Citeseer]{
    \includegraphics[width=7.5cm]{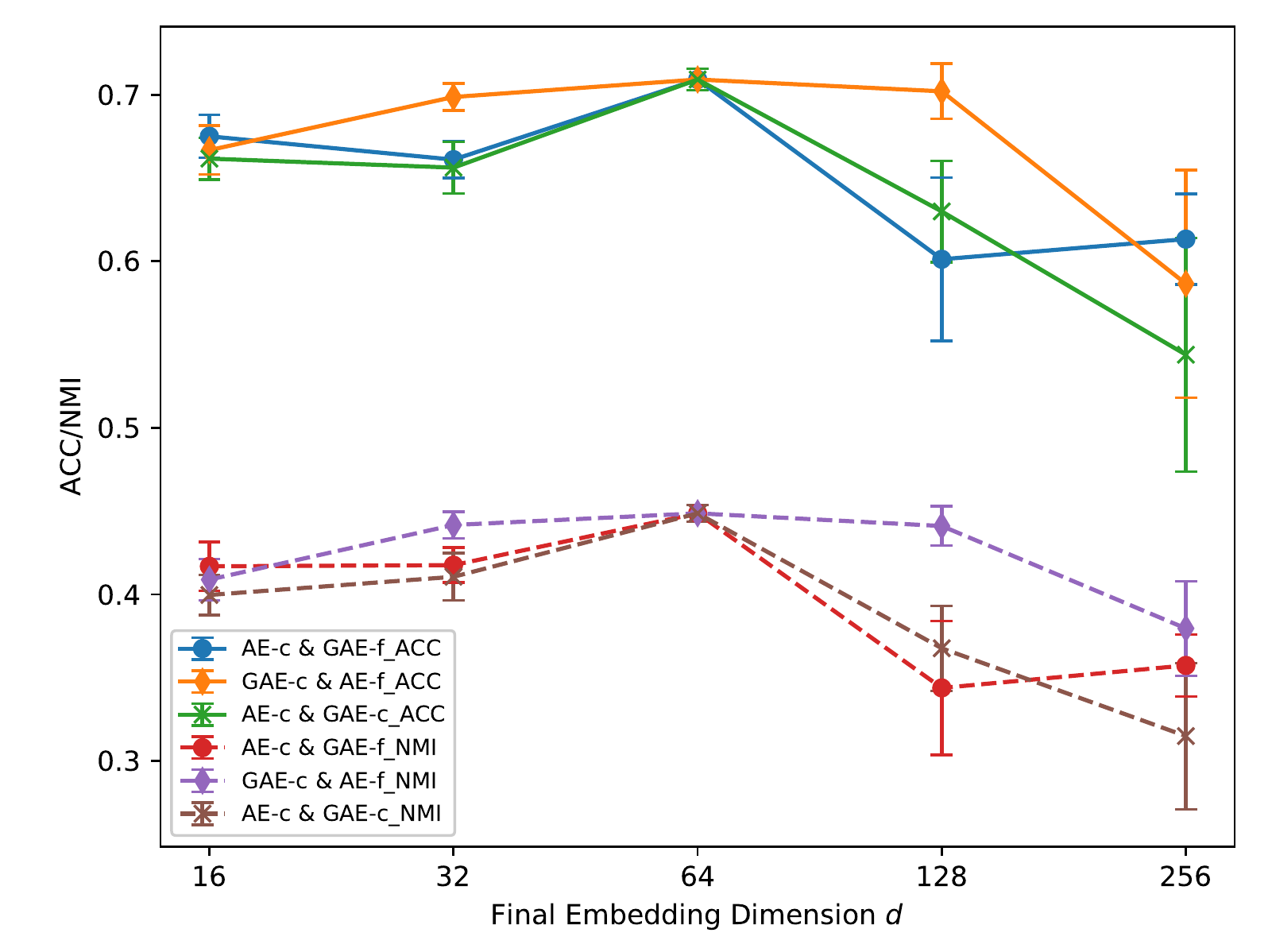}}
    \quad
    \subfloat[ACM]{
    \includegraphics[width=7.5cm]{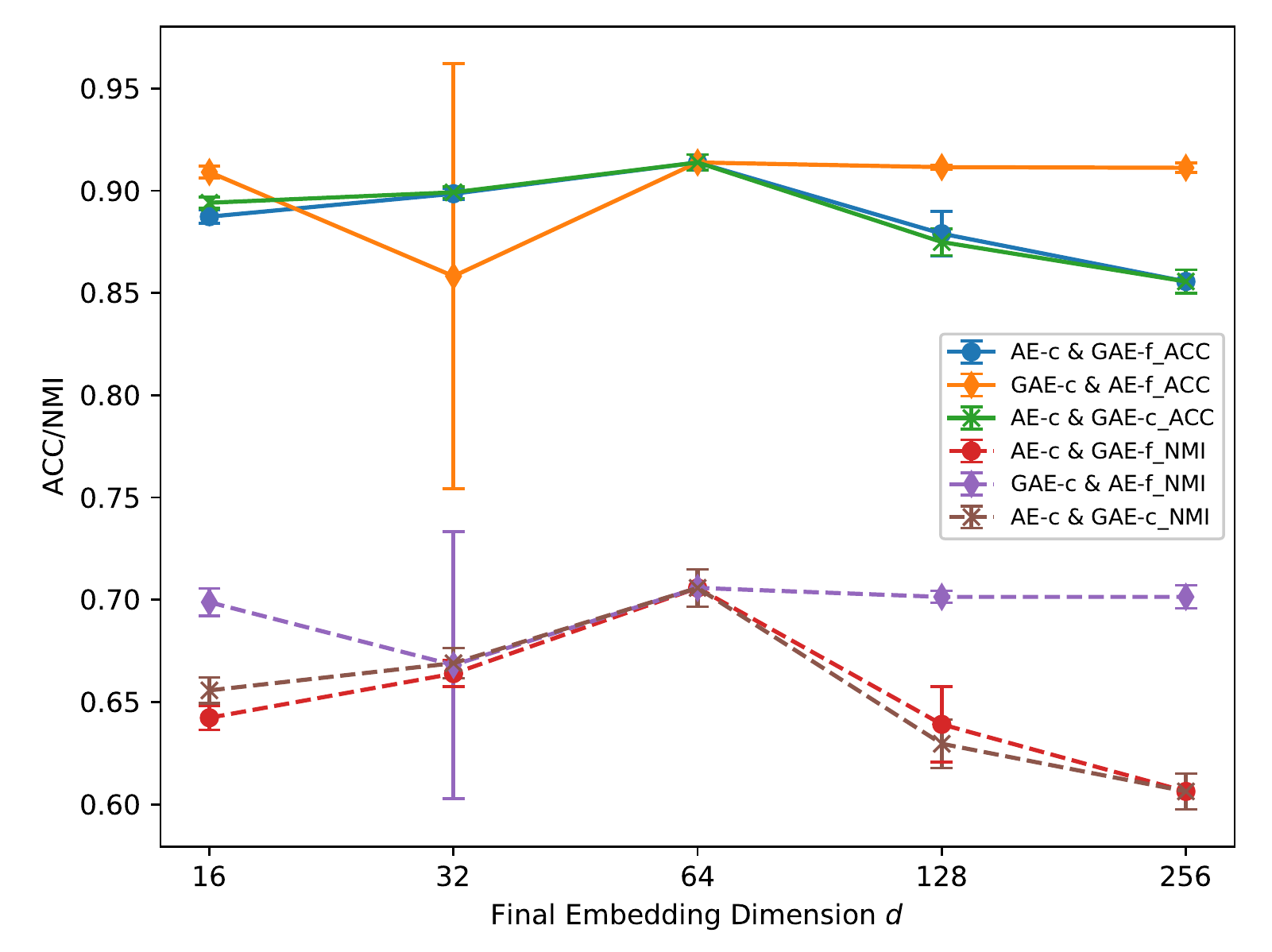}}
    \caption{The effect of the varied final embedding dimension $d$ on Citeseer and ACM networks.}
    \label{fig:dimension}
\end{figure*}

First of all, we examine the effect of the final dimension $d$ of representation vectors in two DEC modules. In detail, we set $d = \{16, 32, 64, 128, 256\}$ and carry out the experiments from three aspects. 
1) Fix the final dimension $d$ of GAE-based module as 64, then change the dimension of the AE-based module; 2) Inversely, fix $d$ of the AE-based module, and change that of the GAE-based module; 3) Change the dimensions of both modules simultaneously, but keep the two dimensions be identical.
Fig.~\ref{fig:dimension} shows the results of Citeseer and ACM as examples, where `AE-c \& GAE-f', `GAE-c \& AE-f', and `AE-c \& GAE-c' correspond to the above three cases. 
For simplicity, we only show the performance of ACC and NMI, and represent them with solid and dashed lines respectively.

From Fig.~\ref{fig:dimension}, we can see that only changing the dimension of the GAE-based module has little impact on the clustering performance in most cases, indicating the GAE-based module is robust to $d$. 
Besides, once we change the final dimension $d$ of representation vectors of AE-based module or both of the two modules, the clustering performance rises when $d$ enlarges from 16 to 64. However, when $d > 64$, the accuracy begins to drop significantly.
This is because a high dimensionality may introduce redundant and noisy information that can harm the clustering performance. 
Therefore, the dimension of representation vectors in the AE-based module for node attributes has a greater impact on the final clustering performance. 
Thus, we choose 64 as the final representation dimension to extract useful information hidden in attributed networks.

\subsubsection{The number and the dimensions of hidden layers}

In this section, we study the impact of the number and the dimensions of hidden layers. 
Since the GAE-based module only utilizes two hidden layers, we focus on testing the performance of different settings for the AE-based module. 
Here we take USPS as an example. The validation process for other datasets is similar.
Table~\ref{tab: hidden layer} presents the clustering results when the number of hidden layers changes from two to five. Once the number of hidden layers is given, we also compare the impact of the different dimensions of the layers on clustering performance. In Table~\ref{tab: hidden layer}, the best performance is highlighted in bold, and the underlined results indicate the sub-optimal results for other settings.

\begin{table*}[ht]
\begin{center}
\caption{The clustering performance under the different number and the different dimensions of hidden layers for USPS.}
\label{tab: hidden layer}
\begin{tabular}{c|cccc}
\hline
hidden   layer               & ACC                     & NMI                     & ARI                     & PWF                     \\
\hline
1) 256-64                       & 0.6339$\pm$0.0164          & 0.6380$\pm$0.0049          & 0.5252$\pm$0.0157          & 0.6032$\pm$0.0243          \\
\underline{2) 512-64}                 & \underline{0.6856$\pm$0.0008}    & \underline{0.6801$\pm$0.0011}    & \underline{0.5770$\pm$0.0013}    & \underline{0.6713$\pm$0.0009}    \\
\cdashline{1-5}
3) 256-256-64                   & 0.6528$\pm$0.0013          & 0.6441$\pm$0.0015          & 0.5325$\pm$0.0016          & 0.6346$\pm$0.0015          \\
4) 512-256-64                   & 0.6733$\pm$0.0012          & 0.6660$\pm$0.0010          & 0.5577$\pm$0.0026          & 0.6590$\pm$0.0013          \\
\underline{5) 512-512-64}             & \underline{0.6836$\pm$0.0027}    & \underline{0.6779$\pm$0.0019}    & \underline{0.5751$\pm$0.0025}    & \underline{0.6679$\pm$0.0029}    \\
\cdashline{1-5}
6) 512-256-256-64               & 0.7239$\pm$0.0010          & 0.7360$\pm$0.0011          & 0.6313$\pm$0.0019          & 0.7141$\pm$0.0010          \\
\underline{7) 512-512-256-64}         & \underline{0.7444$\pm$0.0013}    & \underline{0.7505$\pm$0.0013}    & \underline{0.6574$\pm$0.0024}    & \underline{0.7327$\pm$0.0014}    \\
\cdashline{1-5}
8) 1024-512-256-64              & 0.7321$\pm$0.0011          & 0.7384$\pm$0.0014          & 0.6393$\pm$0.0019          & 0.7195$\pm$0.0011          \\
9) 512-512-512-256-64           & 0.7581$\pm$0.0007          & 0.7748$\pm$0.0014          & 0.6883$\pm$0.0015          & 0.7567$\pm$0.0008          \\
\textbf{10) 1024-512-512-256-64} & \textbf{0.8881$\pm$0.0007} & \textbf{0.8108$\pm$0.0010} & \textbf{0.7846$\pm$0.0013} & \textbf{0.8860$\pm$0.0007}
\\
\hline
\end{tabular}
\end{center}
\end{table*}

According to Table~\ref{tab: hidden layer}, we can draw the conclusion that with the increase of the number of layers, the clustering performance steadily improves.
Among the results in Table~\ref{tab: hidden layer}, cases $\{1),3), 6)\}$ and $\{1), 4), 7), 9), 10)\}$ are obvious examples. 
In addition, when we set the same number of layers, the larger the dimension of the hidden layer, the more significant the clustering performance improves. This means that higher dimensionality can encode more useful information, thus benefiting the joint optimization.
As a result, we use the setting of case 10) for USPS network.

In our experiments, we carefully tune the number of hidden layers and the dimension of each layer for each real-world dataset. We report the result with the best performance among all testing cases. 

\subsubsection{Influence of balance coefficients}

\begin{figure*}[ht]
	\centering
	\subfloat[ACM]{
		\includegraphics[width=5cm]{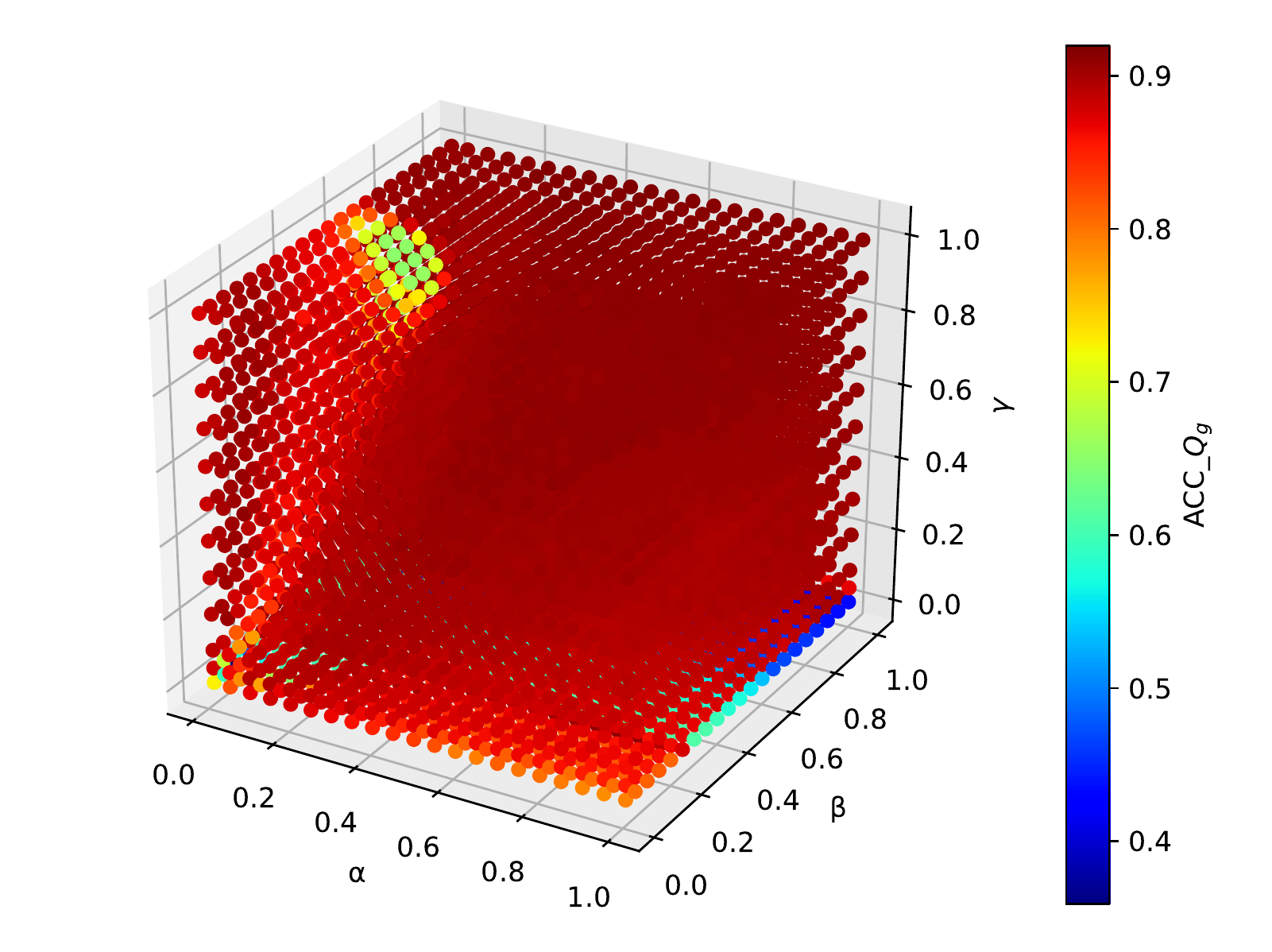}}
	\quad
	\subfloat[DBLP]{
		\includegraphics[width=5cm]{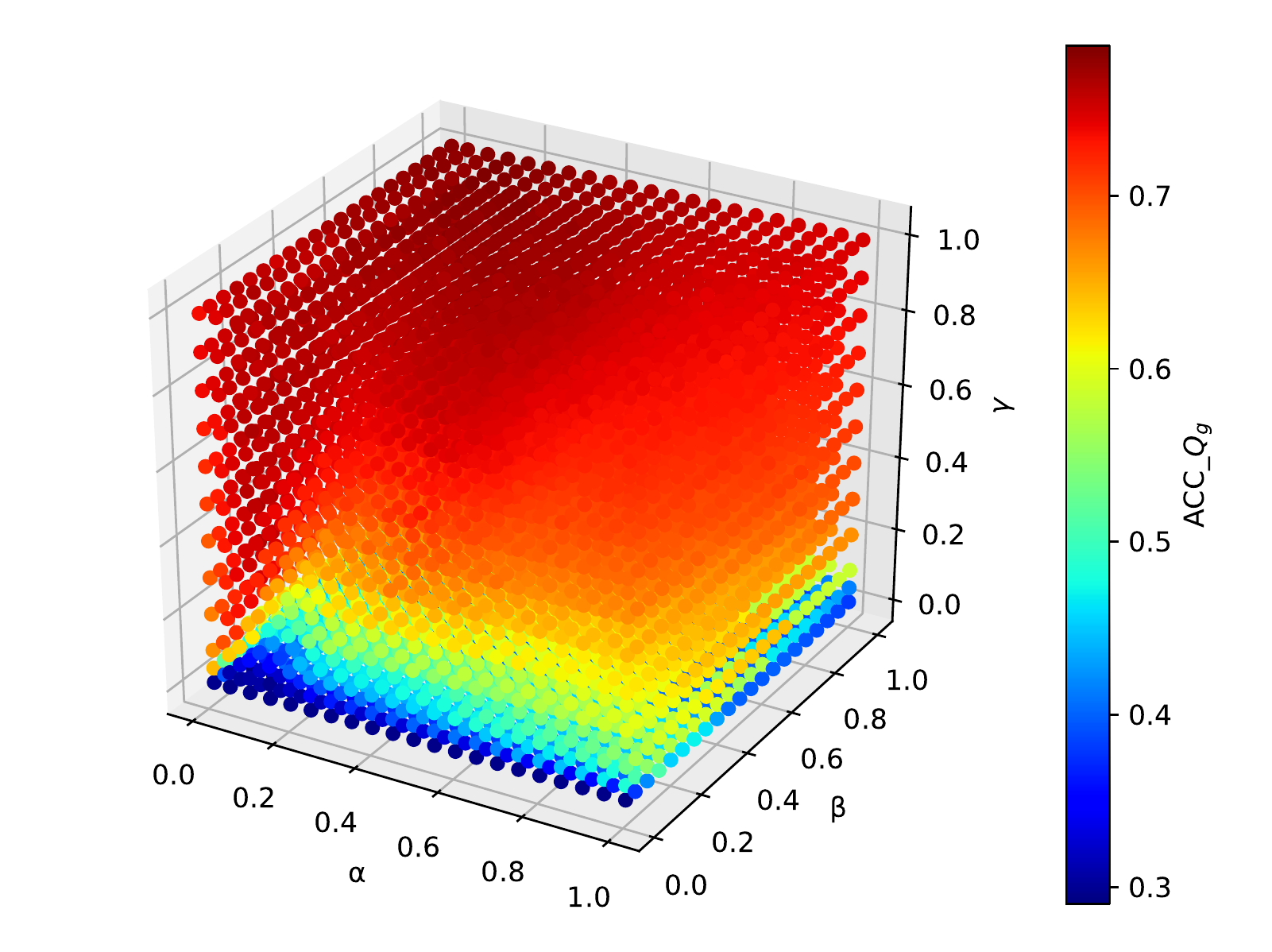}}
	\quad
	\subfloat[USPS]{
		\includegraphics[width=5cm]{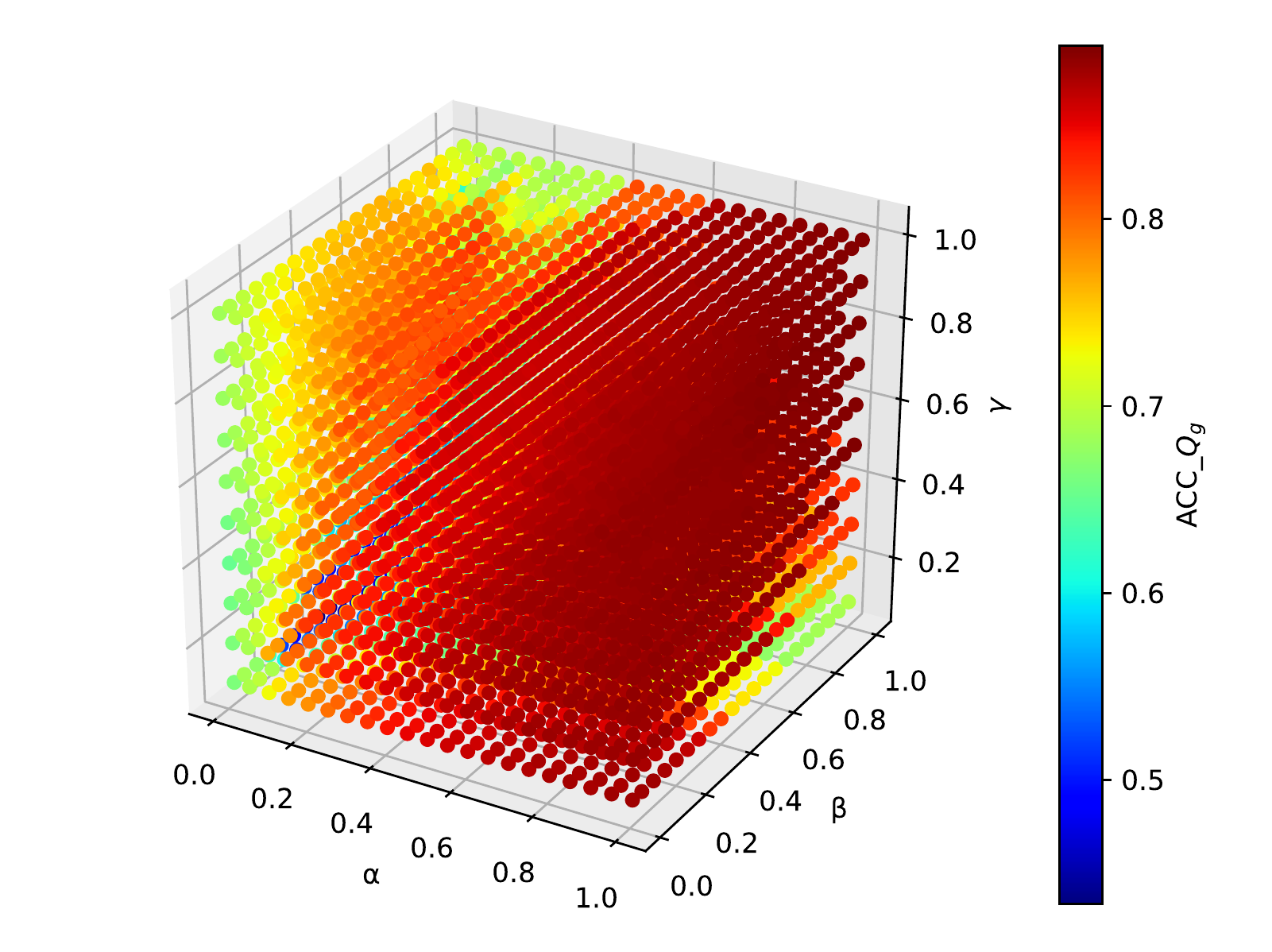}}
	\quad
		\subfloat[ACM]{
		\includegraphics[width=5cm]{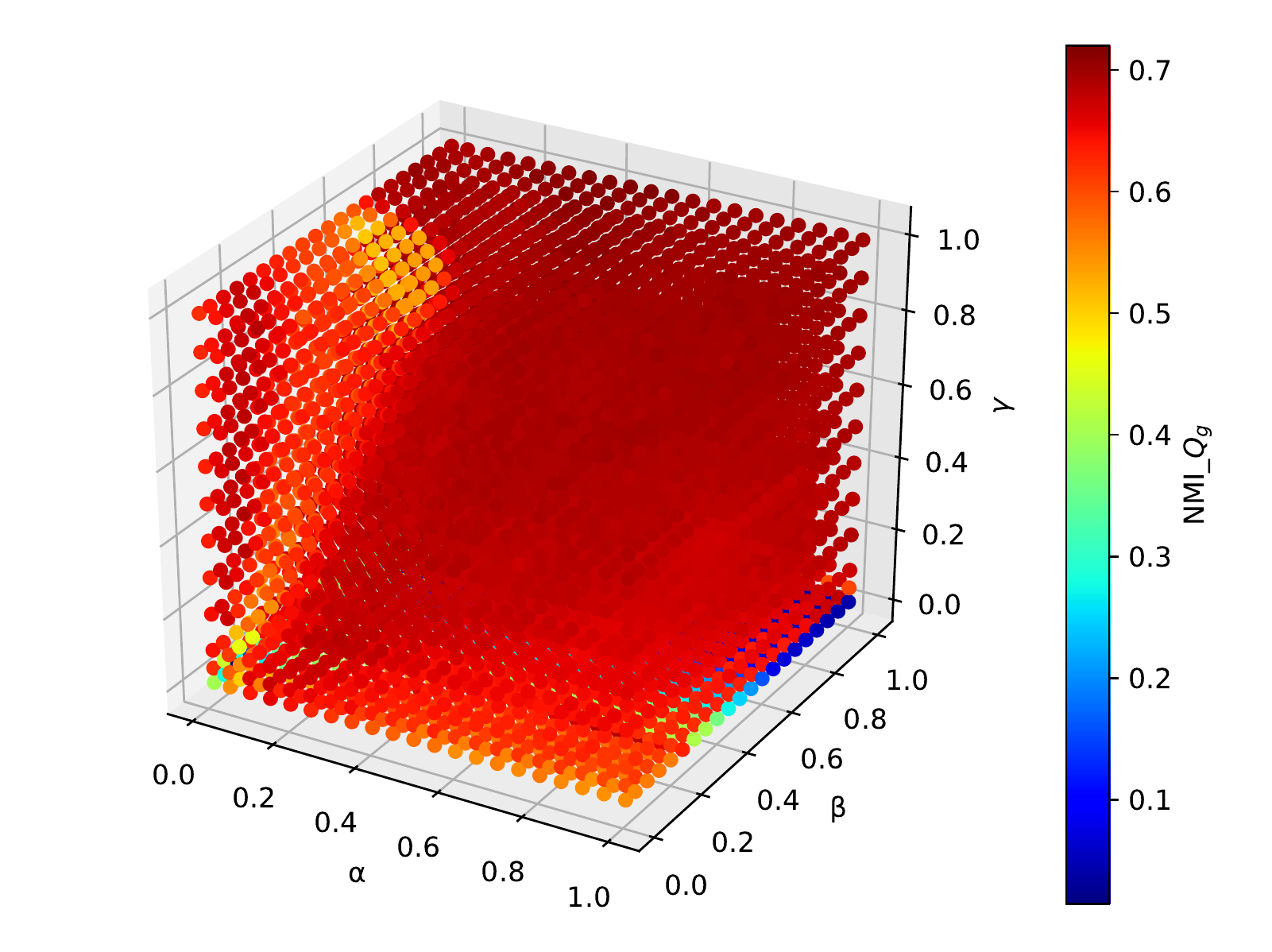}}
	\quad
	\subfloat[DBLP]{
		\includegraphics[width=5cm]{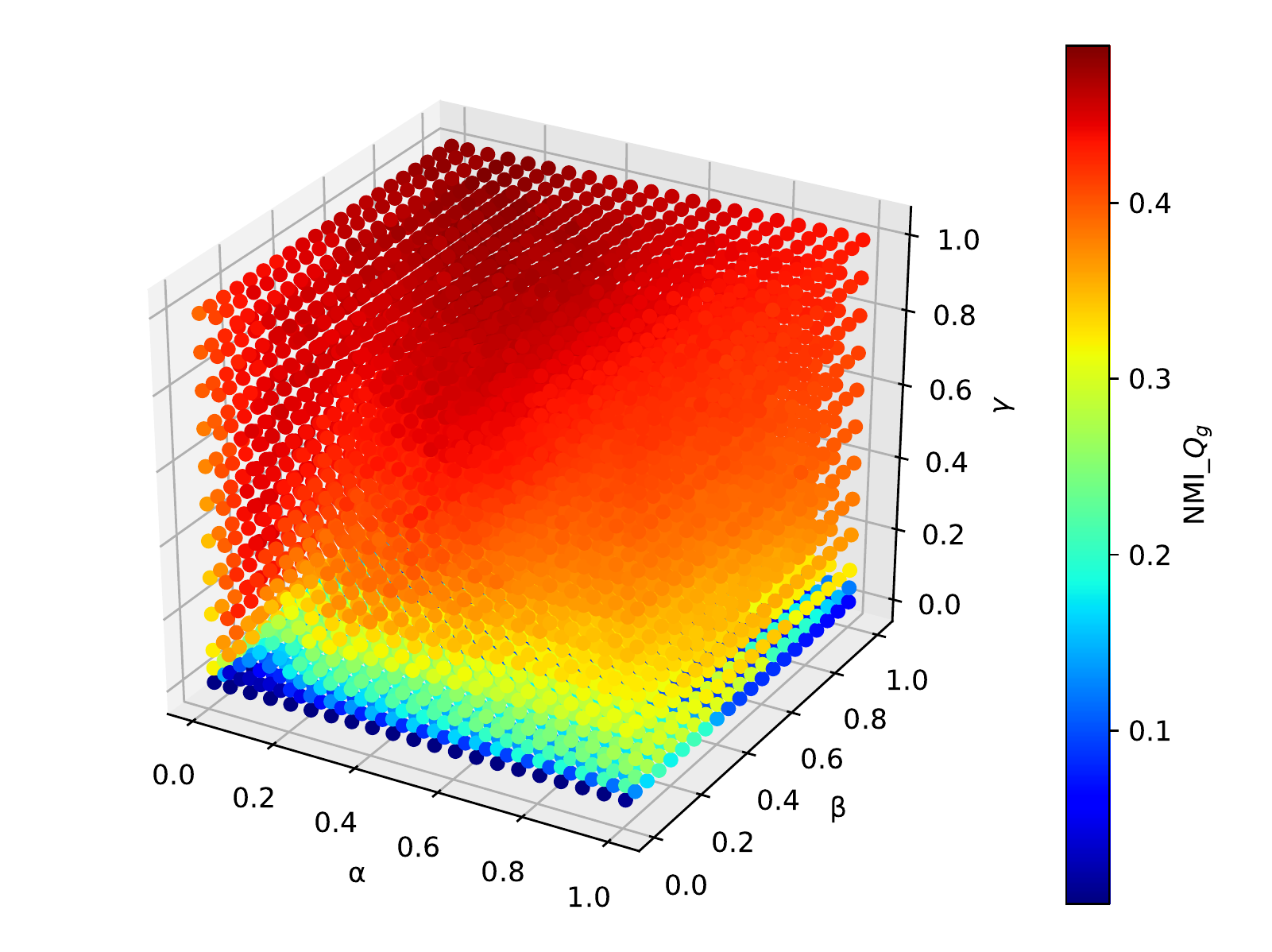}}
	\quad
	\subfloat[USPS]{
		\includegraphics[width=5cm]{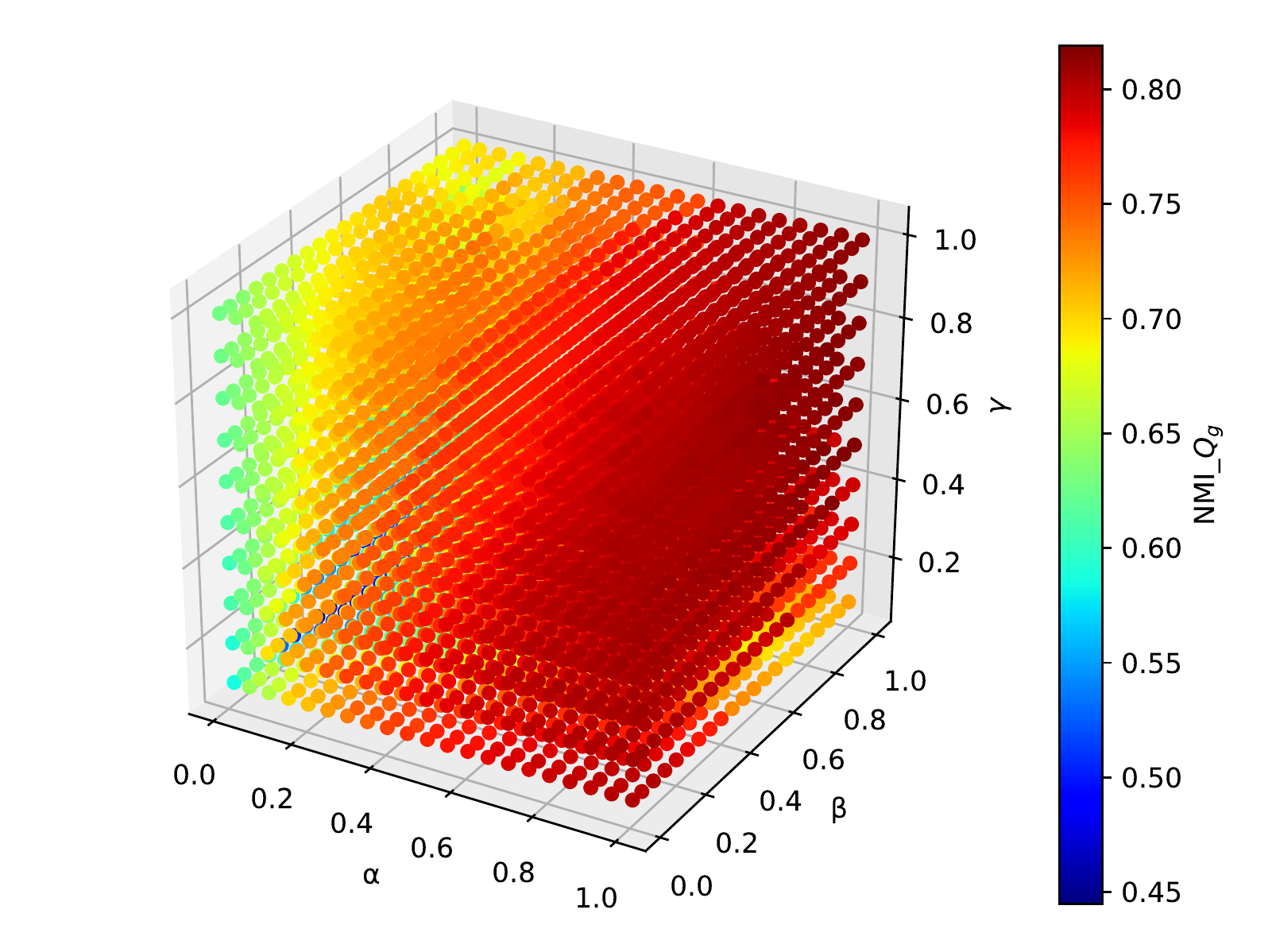}}
	\caption{The influence of the balance coefficients $\alpha$, $\beta$ and $\gamma$ on ACM, DBLP, and USPS networks.
	}
	\label{fig: coefficient}
\end{figure*}

In this section, we verify the tendency of the final clustering performance as the balance coefficient changes (including the weights of clustering loss of two modules $\alpha$ and $\beta$, the coefficient of distribution consistency constraint $\gamma$). We also take three real networks, i.e., ACM, DBLP, and USPS as examples, and present the grid search results on these networks in Fig.~\ref{fig: coefficient} to illustrate the influence of the coefficients, where the figures in the first and the second row correspond to ACC and NMI metrics, respectively. 

From Fig.~\ref{fig: coefficient}, we can see that for different networks, different balance coefficients of each part of the loss have different effects on the overall clustering performance, but in most cases, a variety of coefficient combinations can obtain better clustering structures. More specifically, once we fix $\alpha$ and $\beta$, as $\gamma$ increases, the clustering performance improves gradually. This well verifies the importance of the distribution consistency constraint for combining two DEC modules in DCP-DEC. Meanwhile, although it seems that the choice of $\alpha$ and $\beta$ has relative sensitivity, except for the case of $\alpha = \beta = 0$, a suitable combination will further improve the quality of clusters. This shows again that the introduction of clustering losses in the two DEC modules is useful. At last, the clustering results change little when the value of $\beta$ changes from 0 to 1, regardless of the other two coefficient values ($\gamma > 0$). The performance on ACM network is a clear example. In our experiments, we balance the influence of each part and choose the combinations of coefficients with the best cluster distribution for different networks. In summary, both the KL-based clustering losses formed from network topology and node attributes and the constraint of distribution consistency play an indispensable role during the joint optimization of the whole model.

\subsubsection{Complexity analysis}
In this section, we analyze and compare the efficiency of the proposed model with all baseline methods in terms of their time complexities and computational costs. Following the settings of the original paper, except that DEC and IDEC models adopt the early stopping strategy, we run all other models for 200 epochs and report the training time in Table~\ref{tab:time}.

\begin{table*}[]
	\begin{center}
		\caption{The time complexities and computational costs of compared models.}
		\label{tab:time}
		\resizebox{\linewidth}{!}{\begin{tabular}{cccccccccc}
				\hline
				\multirow{2}{*}{Model} & \multicolumn{1}{c}{\multirow{2}{*}{Time complexity}} & \multicolumn{8}{c}{Computation time (seconds)}                                 \\
				\cline{3-10}
				& \multicolumn{1}{c}{}                                 & CiteSeer & Cora  & PubMed  & ACM   & DBLP  & USPS  & HHAR  & Reuters \\
				\hline
				AE         &     $O(nd^2+tnkd)$        & 38.36    & 26.15 & 177.77  & 29.99 & 36.87 & 82.84 & 92.66 & 102.28  \\
				GAE    &  $O(\vert E \vert d + dn^2 + tnkd)$    & 1.66     & 1.21  & 15.13   & 1.22  & 1.46  & 4.46  & 5.02  & 5.01    \\
				DEC       &       $O(nd^2 + nk)$           & 1.24     & 1.24  & 8.05    & 1.1   & 1.18  & 5.24  & 12.88 & 3.61    \\
				IDEC           &      $O(nd^2 + nk)$        & 1.41     & 1.7   & 10.14   & 1.71  & 3.54  & 8.92  & 33.21 & 10.87   \\
				DAEGC    &   $O(nd^2+\vert E \vert d + dn^2 + nk)$      & 4.29     & 2.98  & 37.88 & 3.26  & 5.46  & 26.15 & 31.04 & 28.83   \\
				SDCN     &    $O(nd^2 + \vert E \vert d + nk)$     & 20.93    & 16.44 & 69.27   & 14.07 & 18.09 & 50.98 & 40.15 & 43.82   \\
				DFCN     &   $O(nd^2 + \vert E \vert d + dn^2 + tnkd)$        & 28.46    & 40.53 & 134.76 & 14.74 & 23.06 & 69.3  & 51.58 & 47.51   \\
				DCP-DEC   &      $O(nd^2 + \vert E \vert d + dn^2 + nk)$ &  10.71     & 8.6   & 52.84   & 7.68  & 9.38  & 33.19 & 26.48 & 31.62   \\
				\hline
		\end{tabular}}
	\end{center}
\end{table*}

In addition, we list the time complexity of each model for a clearer comparison in Table~\ref{tab:time}. Where $n$ and $\vert E \vert$ denote the number of nodes and edges in a graph, $d$ is the maximum dimension of hidden layers, $k$ represents the number of clusters, and $t$ denotes the number of iterations of $K$-means clustering. Taking the DCP-DEC model as an example, the time complexity of the AE-based module is $O(nd^2)$. GAE-based module costs $O(\vert E \vert d + dn^2)$, which includes the GCN encoding and inner product decoding. Meanwhile, computing the cluster distribution takes $O(nk)$. Therefore, the overall time complexity of the proposed DCP-DEC is $O(nd^2 + \vert E \vert d + dn^2 + nk)$. From Table~\ref{tab:time}, we can conclude that although our model does not explicitly pursue computational efficiency during the training, it is also competitive in computational time compared to other algorithms while maintaining the accuracy for clustering tasks.

\section{Conclusion}
In this study, we propose an end-to-end deep embedded clustering model DCP-DEC for attributed networks. 
Our model consists of three parts: AE-based DEC module, GAE-based DEC module, and distribution consistency constraint.
The former two modules make full use of network topology and node attributes to learn node representations and cluster assignments via a self-supervised strategy. 
After that, the distribution consistency constraint characterized by KL divergence is applied to the above two assignments to maintain the latent consistency of two cluster distributions in two DEC modules. 
By minimizing the overall objective function, we can obtain node clusters and simultaneously learn discriminative node representations. 
Extensive experiments on various datasets demonstrate the effectiveness of the proposed model. 
In future work, we will explore more effective and efficient training strategies to jointly optimize the network embedding and node clustering. Meanwhile, it is also worth investigating the different importance of network topology and node attributes for graph clustering, and developing other fusion and distribution measurement strategies.

\section*{Acknowledgment}
This work was supported by the National Key R\&D Program of China (grant numbers 2017YFC1703506, 2018AAA0100302); and National Natural Science Foundation of China (grant numbers 61876016, 61632004).

\bibliographystyle{elsarticle-num}
\bibliography{references}

\end{document}